\documentclass[10pt,twocolumn,letterpaper]{article}

\usepackage[pagenumbers]{wacv}%

\usepackage{array}%
\usepackage{multirow}
\usepackage[ruled,vlined]{algorithm2e}

\usepackage{listings}
\lstdefinelanguage{json}{basicstyle=\ttfamily, showstringspaces=false, breaklines=true}
\definecolor{wacvblue}{rgb}{0.21,0.49,0.74}
\usepackage[pagebackref,breaklinks,colorlinks,allcolors=wacvblue]{hyperref}

\def\wacvPaperID{1283}%

\def\confName{WACV}
\def\confYear{2027}

\title{iARCS: Iterative Agentic RL for Controllable 3D Scene Generation}

\author{Saugat Adhikari$^{1}$* \and
Ashok Prasad Neupane$^{1}$* \and
Pramish Paudel$^{3}$* \and
Ajad Chhatkuli$^{2,3}$ \and
Danda Pani Paudel$^{2,3}$ \\*
\\
$^1$Pulchowk Campus, IOE, Tribhuvan University, Lalitpur, Nepal \\
$^2$NAAMII, Kathmandu, Nepal \\
$^3$INSAIT, Sofia University ``St. Kliment Ohridski'', Bulgaria \\
*Equal contribution \qquad
\href{https://saugat2002.github.io/iarcs-webpage/}{Project page} \quad
\href{https://github.com/thenaivekid/iARCS}{Code} \quad
\href{https://huggingface.co/datasets/Saugat20021/iARCS}{Dataset}
}

\begin{document}
\twocolumn[{%
    \maketitle
    \vspace{-4em}
    \begin{center}
        \centering
        \captionsetup{type=figure}
        \includegraphics[width=\textwidth, trim=0 30 0 28, clip]{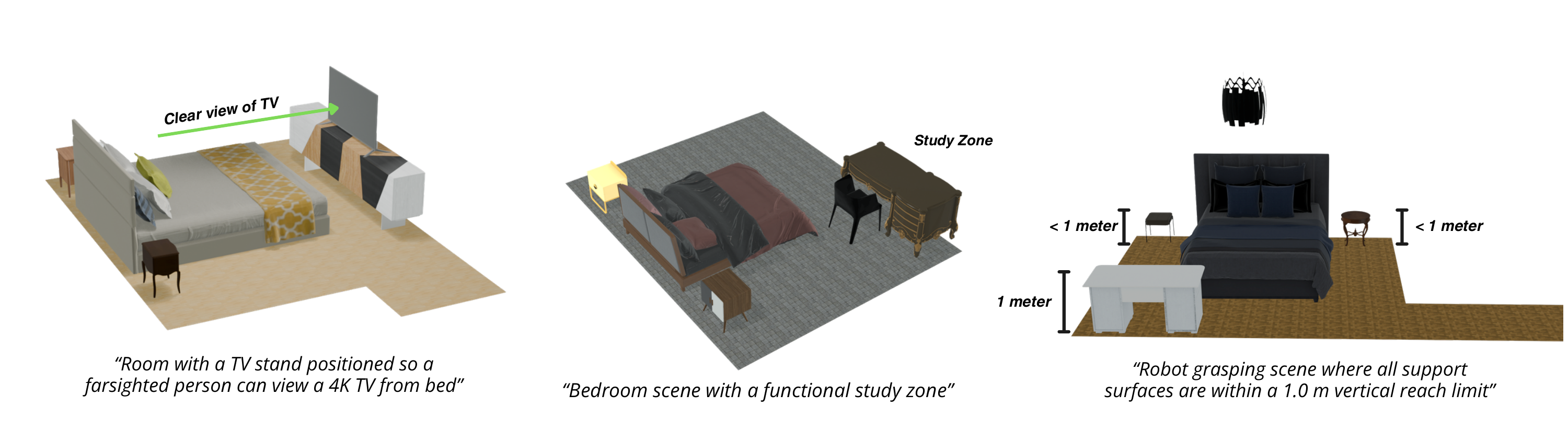}
        \caption{Qualitative results of iARCS under different reward-function specifications. Across tasks, the generated scenes follow the intended reward-driven constraints (e.g., spatial arrangement and functional objectives) while preserving overall scene quality, including physical plausibility and walkable structure.}
        \label{fig:teaser_reward_qualitative}
    \end{center}%
}]

\begin{abstract}
Synthetic 3D scene generation is increasingly used as a data source for computer vision and embodied AI, but existing generators often optimize perceptual realism without reliably satisfying task-critical functional constraints. This mismatch limits the usefulness of synthetic data for downstream training, where accessibility, traversability, and spatial rule compliance are often essential. We present iARCS, an iterative agentic reinforcement learning framework that adapts a pretrained scene generator to natural-language task requirements. iARCS uses a two-phase strategy: universal-reward pretraining to improve physical plausibility and layout quality, followed by task-specific fine-tuning with LLM-generated reward programs that are iteratively refined from training feedback. Experiments show improved constraint fidelity on walkability, reachability, and clearance-focused tasks, effective task-specific constraint optimization, and competitive scene diversity. We further show that data generated by iARCS improves a base generator, supporting its value as a practical synthetic data generation tool rather than only a controllable scene editing method.
\end{abstract}

\section{Introduction}
\label{sec:intro}

Synthetic 3D indoor scene generation is increasingly important for vision and embodied AI, where large-scale diverse training data are essential but expensive to collect and annotate in the real world~\cite{3d_front,roberts2021hypersimphotorealisticsyntheticdataset,deitke2022procthorlargescaleembodiedai,szot2021habitat,kolve2017ai2thor,srivastava2024behavior1k}. Recent generative models~\cite{ho2020ddpm,song2020ddim,lipman2023flowmatching,peebles2023dit,karras2022edm} have significantly improved perceptual realism and distribution matching in scene generation~\cite{atiss,diffuscene,midiffusion}. Yet, many downstream applications demand scenes that satisfy \emph{functional} constraints beyond semantic plausibility~\cite{physcene,szot2021habitat,srivastava2024behavior1k}, \eg, accessibility-aware clearance, controllable traversability for robot testing, and explicit spatial rules over object size, count, and relative placement.
This mismatch creates a practical data bottleneck: benchmark datasets and pre-trained generators rarely provide enough task-critical edge cases~\cite{3d_front,atiss,diffuscene,midiffusion,physcene}. Consequently, generated training data remain weakly aligned to downstream functional use.

Current approaches for controllable 3D scene synthesis fall into three lines: procedural generation \cite{raistrick2024infinigenindoorsphotorealisticindoor,deitke2022procthorlargescaleembodiedai,qi2018humancentricindoorscenesynthesis,chang2015text3dscenegeneration}, LLM-augmented construction \cite{yang2024holodecklanguageguidedgeneration,bian2025holodeck20visionlanguageguided3d,aguinakang2024openuniverseindoorscenegeneration,xia2026sage,Littlefair_2025,yang2025optiscenellmdrivenindoorscene,deng2025globallocaltreesearchvlms} and data-driven generative modeling \cite{atiss,diffuscene,midiffusion,lin2024instructsceneinstructiondriven3dindoor,zhai2023commonscenesgeneratingcommonsense3d,dhamo2021graphto3dendtoendgenerationmanipulation,wang2023rearrangeindoorsceneshumanrobot,ran2025directlayoutdirectnumericallayout,zhou2024gala3dtextto3dcomplexscene,fang2025spatialgenlayoutguided3d,fang2025mvroomcontrollable3dindoor,su2025chordgenerationcollisionfreehousescale}. Procedural methods provide scalable variation \cite{deitke2022procthorlargescaleembodiedai,raistrick2024infinigenindoorsphotorealisticindoor}, but their hand-crafted rules make fine-grained task semantics hard to specify and verify consistently. LLM-based pipelines improve prompt-level flexibility by converting language into scene graphs or editing constraints \cite{yang2024holodecklanguageguidedgeneration,lin2024instructsceneinstructiondriven3dindoor}, yet reliability still depends on LLM scene composition ability, with weak guarantees under complex multi-constraint settings \cite{yang2024holodecklanguageguidedgeneration,lin2024instructsceneinstructiondriven3dindoor}. Data-driven models such as ATISS~\cite{atiss}, DiffuScene~\cite{diffuscene}, MiDiffusion~\cite{midiffusion}, and PhyScene~\cite{physcene} match training distributions and improve plausibility, but still struggle to satisfy user-defined functional constraints.

Text conditioning alone is usually insufficient for strict functional scene requirements: language-guided pipelines improve instruction following, but they do not reliably enforce precise geometric constraints and typically lack explicit verification during generation~\cite{lin2024instructsceneinstructiondriven3dindoor,layoutvlm,yang2024holodecklanguageguidedgeneration}. Reward-guided optimization improves controllability, yet existing practice still depends on manually engineered objectives and substantial reward-design effort, which limits clean scaling to arbitrary new constraints~\cite{physcene,ma2024eurekahumanlevelrewarddesign,black2023trainingdiffusionmodelsrl,pfaff2025steerablescenegenerationpost}. On the other hand, recent agentic reward frameworks~\cite{ma2024eurekahumanlevelrewarddesign,text2reward,uncertainty_reward,language2rewards,agentic_rl_survey} have made rapid progress in several applications. However, reward functions and codes in scene synthesis remain largely static~\cite{metaspatial,sceneweaver,layout_r1}.

We formulate synthetic scene generation as the post-training adaptation of a pretrained scene prior to satisfy user-defined functional constraints while preserving realism and diversity. This view aligns with recent post-training and RL-based control of generative models, which optimize task-specific objectives on top of pretrained diffusion priors rather than rebuilding generators from scratch~\cite{black2023trainingdiffusionmodelsrl,pfaff2025steerablescenegenerationpost}.
In order to realize this goal, we propose a two-phase agentic reinforcement learning (RL) framework for post-training scene generators called iARCS. In \textbf{Phase 1}, we optimize universal rewards (\eg, physical plausibility and design priors) to strengthen baseline scene quality, and in \textbf{Phase 2}, an LLM agent translates user prompts into task-specific reward programs and iteratively refines them using training feedback and reward evolution memory. We fine-tune the diffusion generator with Denoising Diffusion Policy Optimization (DDPO)~\cite{black2023trainingdiffusionmodelsrl} to optimize non-differentiable objectives (e.g., interpenetration, boundary violation, and the task constraint) while regularizing against distribution collapse.

\begin{figure}[t]
\vspace{-1em}
\centering
\includegraphics[width=\columnwidth]{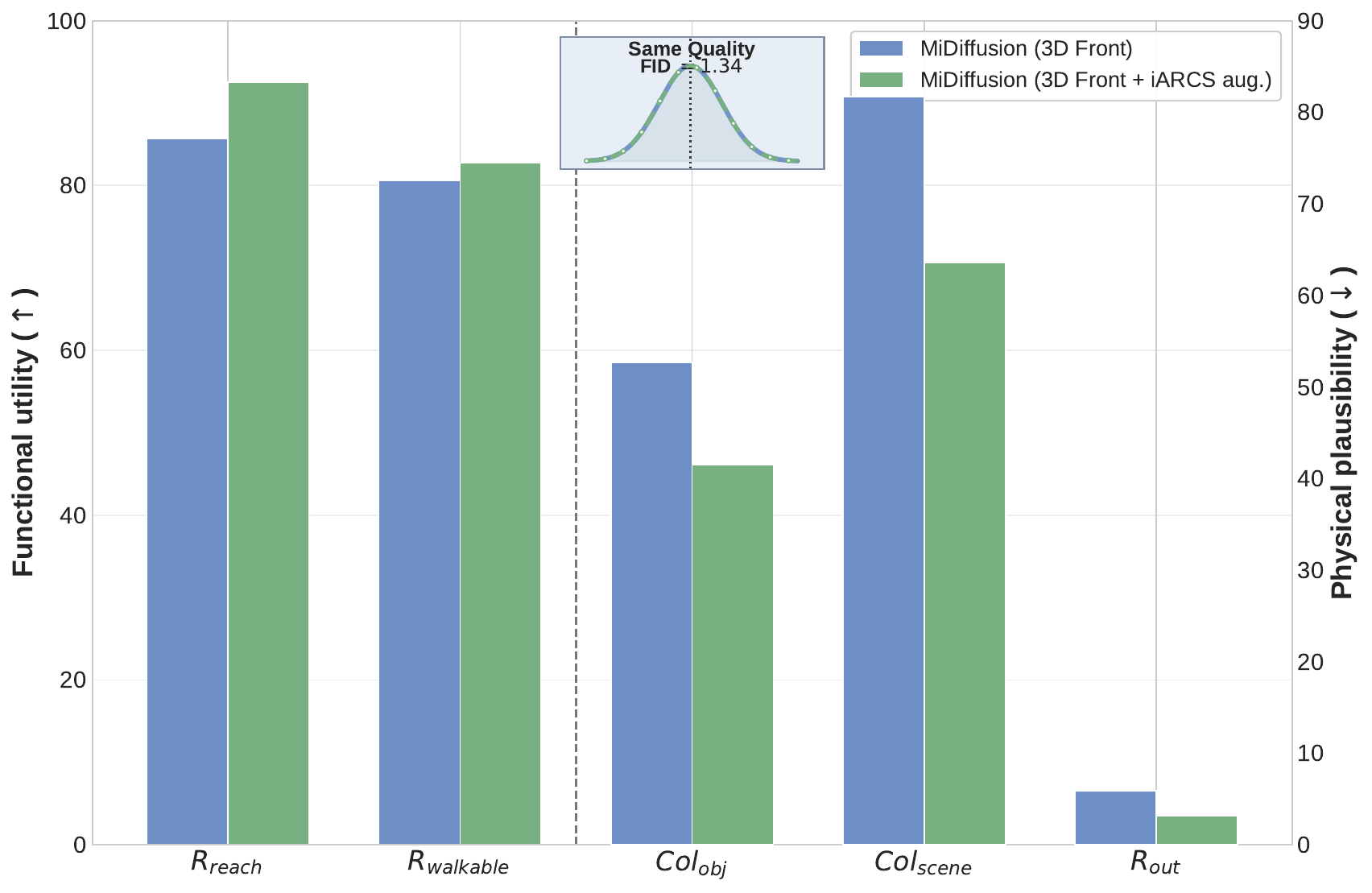}
\caption{\textbf{MiDiffusion trained with iARCS augmentation vs. MiDiffusion trained on 3D-FRONT only.} We compare MiDiffusion trained on 3D-FRONT against MiDiffusion trained on 3D-FRONT + iARCS-generated synthetic data. The augmented setting improves functional utility ($R_{reach}$, $R_{walkable}$) and physical plausibility ($Col_{obj}$, $Col_{scene}$, $R_{out}$) across all reported metrics, while matched FID ($1.34$) indicates preserved generative quality and diversity.}
\label{fig:quantitative_improvement_midiffusion}
\end{figure}

Further, we propose the iterative agentic loop for custom task specific reward generation so that we can control generated scenes without explicit handcrafted reward design utilizing reward evolution memory and reward reflection.
This design addresses both failure modes: it retains the realism and diversity of learned scene distributions while enabling explicit enforcement of functional constraints through reward-driven adaptation. Empirically, the two-phase optimization schedule is substantially more effective than single-phase training with joint optimization. As summarized in Fig.~\ref{fig:quantitative_improvement_midiffusion}, we find that iARCS functions as an effective synthetic-data engine: training MiDiffusion with iARCS-generated synthetic data improves a strong state-of-the-art indoor scene generator while preserving diversity and realism.

Our contributions are as follows:
\begin{itemize}
    \item We propose iARCS, an iterative agentic RL framework for controllable 3D scene generation from natural-language constraints.

    \item We introduce a two-phase training strategy (universal-reward pretraining + task-specific fine-tuning) that improves physical plausibility and functional utility.

    \item We show effective task-specific constraint optimization and that iARCS-generated data improves a base generator while maintaining competitive diversity.
\end{itemize}

\section{Related Works}
\label{sec:related}

\paragraph{Generative Priors for 3D Scenes.}
Recent approaches to 3D indoor scene synthesis leverage deep generative models to capture complex geometric and semantic distributions, including primitive-based scene decomposition and learned point-set representations~\cite{fedele2025superdec3dscenedecomposition,qi2017pointnetdeeplearningpoint}. ATISS~\cite{atiss} and SceneFormer~\cite{wang2021sceneformerindoorscenegeneration} use autoregressive transformers to sequence object placements, while diffusion, layout-guided, and arrangement-based methods~\cite{diffuscene,midiffusion,physcene,wei2023legonetlearningregularrearrangements,yang2025scenecraftlayoutguided3dscene} map latent noise or layout conditions to structured layouts $S$. These architectures typically learn a distribution $p(S|F)$ conditioned on a floor boundary $F$. However, downstream applications often require scenes to satisfy specific functional or spatial rules $C$, as highlighted by controllable, language-guided, or retrieval-augmented scene-generation methods~\cite{physcene,lin2024instructsceneinstructiondriven3dindoor,yang2024holodecklanguageguidedgeneration,bian2025holodeck20visionlanguageguided3d,feng2023layoutgpt,feng2025casagptcuboidarrangementscene,sun20243dgptprocedural3dmodeling,bucher2025respace}. Sampling from the resulting constrained posterior $p(S|F, C)$ is non-trivial because these functional dependencies are rarely represented in the training data, a challenge also reflected in scene-graph and commonsense scene generation methods~\cite{wang2021sceneformerindoorscenegeneration,zhai2023commonscenesgeneratingcommonsense3d,dhamo2021graphto3dendtoendgenerationmanipulation,feng2023layoutgpt,bucher2025respace}. Naive approaches like rejection sampling from the prior $p(S|F)$ are computationally inefficient and often fail to produce diverse scenes~\cite{pfaff2025steerablescenegenerationpost,fan2023dpokreinforcementlearningfinetuning}. Consequently, explicit guidance or optimization is required to shift the generative prior toward these narrow, task-critical regions of the scene space~\cite{ho2022classifierfreediffusionguidance,physcene,singhal2025generalframeworkinferencetimescaling,black2023trainingdiffusionmodelsrl}.

Table~\ref{tab:related_comparison} summarizes the key distinction between representative scene-generation paradigms. Unlike LLM/VLM-composed methods, iARCS uses language only to specify rewards rather than place geometry, and unlike differentiable guidance methods, it optimizes non-differentiable constraints through RL fine-tuning of a learned scene prior.

\begin{table}[t]
\centering
\footnotesize
\setlength{\tabcolsep}{2.5pt}
\renewcommand{\arraystretch}{1.12}
\resizebox{\columnwidth}{!}{%
\begin{tabular}{>{\raggedright\arraybackslash}p{0.25\columnwidth}p{0.12\columnwidth}p{0.22\columnwidth}p{0.15\columnwidth}p{0.18\columnwidth}}
\hline
\textbf{Method} & \textbf{Scene prior} & \textbf{Geometry placement} & \textbf{User constraints} & \textbf{Non-diff. constraints} \\
\hline
SAGE~\cite{xia2026sage}, Holodeck~\cite{yang2024holodecklanguageguidedgeneration} & No & LLM/VLM & Yes & Yes \\
\hline
ATISS~\cite{atiss}, SceneFormer~\cite{wang2021sceneformerindoorscenegeneration}, DiffuScene~\cite{diffuscene}, MiDiffusion~\cite{midiffusion} & Yes & Learned model & No & No \\
\hline
PhyScene~\cite{physcene} & Yes & Model + guidance & Yes & Diff. only \\
\hline
\textbf{iARCS (ours)} & \textbf{Yes} & \textbf{Learned diffusion model} & \textbf{Yes} & \textbf{Yes; RL rewards} \\
\hline
\end{tabular}
}
\caption{Comparison of representative 3D scene generation methods. iARCS preserves a learned scene prior while using LLM-generated rewards, rather than LLM/VLM geometry placement, to optimize user-specified and non-differentiable constraints.}
\label{tab:related_comparison}
\end{table}

\paragraph{Guidance and Constraint Satisfaction in Diffusion.} To enforce specific constraints during generation, prior literature has explored various inference-time and train-time guidance mechanisms. At inference time, Classifier-Free Guidance (CFG)~\cite{ho2022classifierfreediffusionguidance} is standard for aligning outputs with text embeddings, while PhyScene~\cite{physcene} applies training-time and test-time gradient guidance utilizing differentiable guidance functions to enforce physical rules; recent text-driven layout/scene synthesis methods further highlight the need for stronger controllability under natural-language constraints~\cite{lin2024instructsceneinstructiondriven3dindoor,yang2024holodecklanguageguidedgeneration,bian2025holodeck20visionlanguageguided3d}. Inference-time scaling or alignment methods~\cite{singhal2025generalframeworkinferencetimescaling,ma2025inferencetimescalingdiffusionmodels,jajal2025inferencetimealignmentdiffusionmodels} provide another route; for instance, Steerable Scene Generation~\cite{pfaff2025steerablescenegenerationpost} employs Monte Carlo Tree Search (MCTS) to navigate the generative trajectory toward valid states. However, inference-time guidance faces fundamental limitations: gradient-based methods strictly require all user-defined constraints to be differentiable, and search-based methods incur prohibitive latency. Agentic generation-time systems such as SAGE~\cite{xia2026sage} face the same recurring cost: planning and correction run per generated scene. iARCS instead pays the agentic cost once, during post-training, and leaves an ordinary diffusion generator that samples with no LLM in the loop; this also means the adaptation improves the \emph{generator}, so its samples are reusable as training data rather than being one-off outputs. In order to bypass these issues, train-time optimization via reinforcement learning has emerged as an alternative, utilizing policy optimization~\cite{black2023trainingdiffusionmodelsrl,fan2023dpokreinforcementlearningfinetuning,pfaff2025steerablescenegenerationpost,zhang2024largescalereinforcementlearningdiffusion,nabla_r2d3,dreamreward,dreamdpo,dreamcs} to fine-tune the diffusion model directly toward specific, non-differentiable objectives.

\paragraph{Agentic RL Formulation.} Applying reinforcement learning to diffusion models requires robust, dense reward signals. Traditional constraint-satisfaction pipelines rely heavily on hand-crafted, mathematically rigid reward functions, which demand significant engineering effort and fail to scale across arbitrary, user-specified spatial rules. Recently, Eureka \cite{ma2024eurekahumanlevelrewarddesign} used a Large Language Model (LLM) as an automated reward generator within an agentic loop, writing executable reward programs that surpass human-engineered functions. While Eureka \cite{ma2024eurekahumanlevelrewarddesign} focuses on iterative curriculum learning for robotic control policies in physical simulators, iARCS adapts this paradigm to the generative space. By transitioning from rigid manual objectives to dynamic, LLM-generated reward programs, our two-phase RL framework enables the post-training adaptation of diffusion priors to complex, non-differentiable functional constraints~\cite{ma2024eurekahumanlevelrewarddesign,black2023trainingdiffusionmodelsrl}.

\section{Method}

Our method is a two-phase framework. In Phase 1, we use reinforcement learning (RL) to correct biases in a diffusion model trained on the base data distribution by optimizing for physical plausibility: collision avoidance and in-bound placement. In Phase 2, given a user prompt, an LLM module generates task-specific reward functions on top of Phase 1 constraints, and the model is further optimized through iterative reward reflection. In our experiments, we use Gemini~\cite{geminiteam2025geminifamilyhighlycapable} model as LLM function generator and perform reward reflections every 10 RL stages, where one RL stage is a single rollout-generation and policy-update cycle.

\begin{figure*}[h]
\vspace{-2.5em}
\centering
\includegraphics[width=\textwidth]{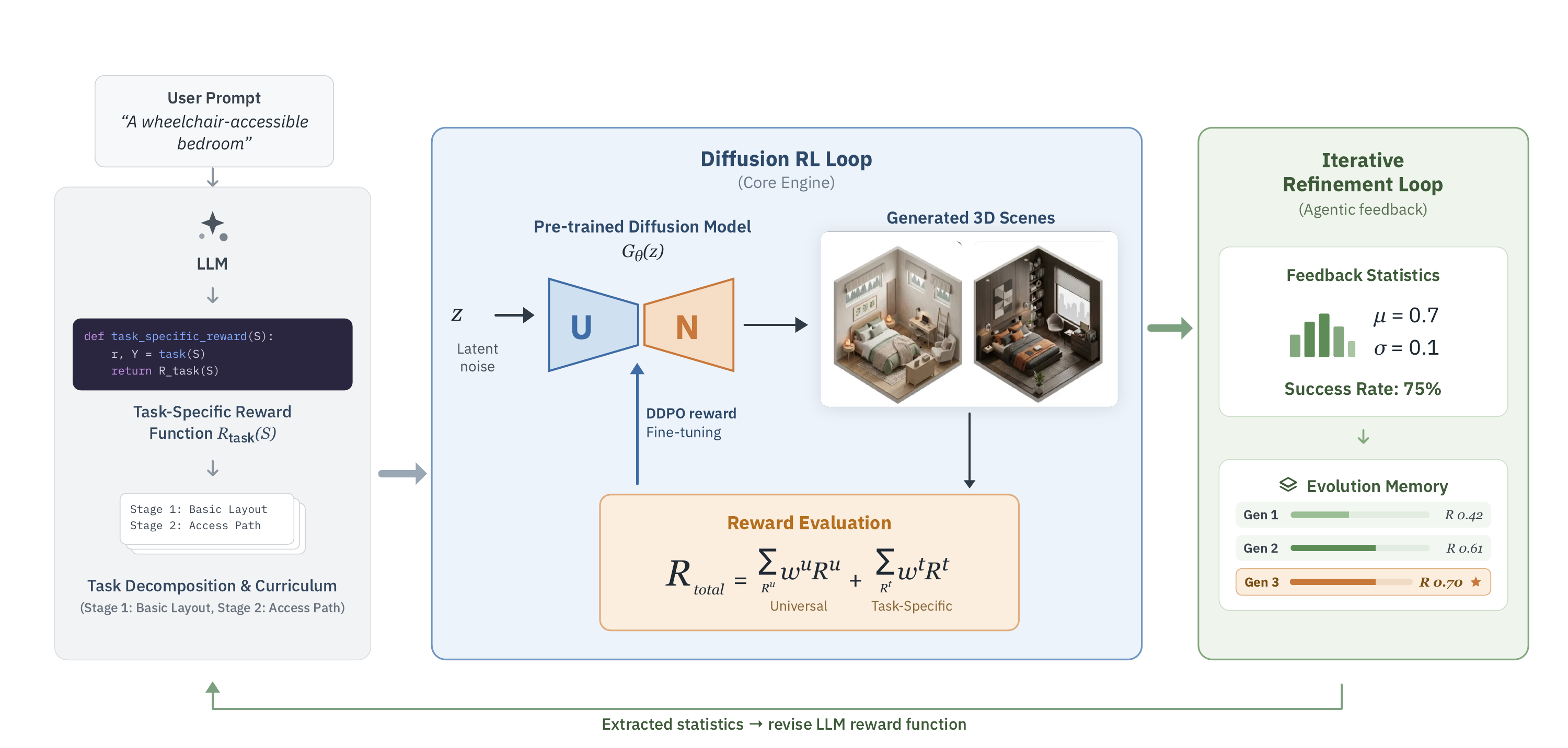}
\caption{\textbf{iARCS overview: agentic reward synthesis with diffusion RL and reflection.} Given a user prompt, an LLM agent performs (1) reasoning, (2) constraint decomposition, and (3) executable reward-program generation. The generated task reward is combined with universal rewards to form a composite objective, which is optimized by DDPO in the diffusion RL loop to fine-tune a pretrained scene generator. A reward-reflection module monitors reward statistics from generated scenes, revises reward code/weights, and feeds updates back to the agent, enabling iterative improvement in constraint satisfaction while preserving scene realism and diversity.}
\label{fig:overview}
\vspace{-3mm}
\end{figure*}

\subsection{Problem Formulation}
\noindent\textbf{Scene Representation.} Following prior work~\cite{midiffusion, atiss}, we define a 3D scene as an unordered set of $N$ objects, $S = \{o_1, o_2, \dots, o_N\}$. Each object $o_i$ is parameterized by its continuous geometric attributes and discrete semantic category:
\begin{equation}
o_i = (t_i,\, s_i,\, \cos(\theta_i),\, \sin(\theta_i),\, c_i)
\end{equation}
where $t_i \in \mathbb{R}^3$ is the centroid translation, $s_i \in \mathbb{R}^3$ represents the 3D dimensions (width, height, depth), $\theta_i \in [0, 2\pi)$ is the orientation around the vertical axis, and $c_i \in \mathbb{R}^K$ is the one-hot encoded object category for $K$ classes.

Let $F$ denote a given floor boundary represented as image feature~\cite{atiss} and $C$ be a user-specified functional constraint expressed in natural language. We treat the 3D scene generation process as a floor-plan-conditioned policy $\pi_\theta(x|F)$ parameterized by $\theta$, which generates a scene $S$ conditioned on the floor plan $F$. For each constraint $C$, iARCS adapts the pretrained parameters to a task-specific LoRA policy with parameters $\theta_C$. Our primary objective is to find the adapted parameters $\theta_C$ that maximize the expected composite reward:
\begin{equation}
\label{eq:objective}
\theta_C = \arg\max_{\theta}\; \mathbb{E}_{x \sim \pi_\theta(x \mid F)}\left[R_{\text{total}}(S, C, F)\right]
\end{equation}
where the total reward $R_{\text{total}}$ evaluates the fundamental physical plausibility and functionality of the generated scene and its adherence to the user's high-level functional constraint $C$. Thus, the final generator used for a task is $\pi_{\theta_C}(x|F)$; we do not assume a single zero-shot multi-task policy conditioned directly on $C$.

\subsection{Diffusion Formulation and Policy Optimization}
We model scene generation with a conditional diffusion process~\cite{ho2020ddpm,song2020ddim}. Given clean scene parameters $x_0$ and condition $F$, the forward noising process is
\begin{equation}
q(x_t\mid x_{t-1}) = \mathcal{N}(\sqrt{1-\beta_t}x_{t-1}, \beta_t I),\quad t=1,\dots,T,
\end{equation}
and the learned reverse policy $\pi_\theta$ denoises
\begin{equation}
\pi_\theta(x_{t-1}\mid x_t,F)=\mathcal{N}(\mu_\theta(x_t,t,F),\Sigma_t).
\end{equation}
To adapt this pretrained prior to a functional constraint $C$, we treat denoising as a finite-horizon Markov Decision Process and optimize with DDPO~\cite{black2023trainingdiffusionmodelsrl}. For terminal reward $R_{\text{total}}(x_0,C,F)$, the task-specific adaptation objective is
\begin{equation}
\label{eq:ddpo}
\theta_C = \arg\max_\theta\;J_C(\theta)=\mathbb{E}_{x_{0:T}\sim \pi_\theta(\cdot\mid F)}\left[R_{\text{total}}(x_0,C,F)\right],
\end{equation}
with policy-gradient updates on denoising transitions, yielding the adapted policy $\pi_{\theta_C}$. The denoising state carries both the geometric attributes and the category channels of every object, and the DDIM log-probability is taken over all of them, so the policy gradient reaches object categories as well as continuous layout. Optimizing eq.~\eqref{eq:ddpo} requires careful regularization in order to preserve the pretrained model prior while improving constraint satisfaction.

\subsection{Framework Overview}
In order to achieve the objective in eq.~\eqref{eq:objective} with the formulation in eq.~\eqref{eq:ddpo}, without suffering from catastrophic forgetting or structural degradation, we propose the iARCS framework, as illustrated in Fig.~\ref{fig:overview}. The framework operates in two distinct optimization phases. Throughout, \emph{phase} refers to these two optimization phases, and \emph{RL stage} to a single rollout-generation and policy-update cycle:
\begin{itemize}
    \item \textbf{Phase 1: Base Model Refinement with Universal Rewards.} Pretrained generative models often internalize dataset flaws such as object collisions. To rectify this, we first optimize the base policy $\pi_\theta$ using a set of universal, rule-based rewards ($R^u$). These enforce basic physical laws and structural integrity.
    \item \textbf{Phase 2: Functional Adaptation via Agentic Feedback.} Once a physically plausible baseline is established, we introduce the user constraint $C$. As shown in Fig.~\ref{fig:overview}, an LLM agent translates the natural language prompt into an executable reward program ($R^t$). The model is then optimized using the composite reward $R_{\text{total}}$, with support from an iterative reward-reflection loop that continuously improves the reward design, prevents reward hacking, and ensures stable learning.
\end{itemize}

\subsection{Agentic Reward Synthesis}
Illustrated in the \textbf{Agent Process} module in our pipeline in Fig.~\ref{fig:overview}, a Large Language Model (LLM) agent translates the user prompt into an executable reward program, denoted as $R^t$, following recent language-driven scene reasoning and programmatic control paradigms~\cite{ma2024eurekahumanlevelrewarddesign,feng2023layoutgpt,yang2024holodecklanguageguidedgeneration,bucher2025respace}. This synthesis is performed inside an iterative loop with \textbf{reward evolutionary memory}, which stores the reward functions generated in previous iterations together with the training progress and reward statistics observed at each stage. At every reflection step, the LLM uses this memory to reason about failure modes, revise the constraint decomposition, and write an improved reward function for the next RL stage. Each reward-generation or reward-reflection step follows a three-step procedure:
\begin{enumerate}
    \item \textbf{Reasoning:} The agent identifies critical, restrictive constraints and spatial dependencies implied by the prompt. For example, given the prompt ``narrow bedroom with 0.2m walking path", the agent reasons: \textit{``A 0.2m walking path is a critical, restrictive constraint. Furniture must be pushed against the walls to ensure a continuous 0.2m wide aisle from the door to the bed."}
    \item \textbf{Decomposition:} Abstract constraints are broken down into atomic, measurable geometric checks. Following the previous example, this decomposes into: \textit{Walking Space = 0.2m (Check), Furniture Alignment (against the walls) and Path Clearance (continuous aisle).}
    \item \textbf{Reward Function:} The agent outputs executable Python code (e.g., {\small \texttt{def robot\_3DFRONT\_walkability\_reward(scene): \dots}}) that computes a scalar reward based on the generated scene and the decomposed metrics.
\end{enumerate}
This memory-guided reflection loop enables the agent to avoid repeatedly generating ineffective rewards, adapt the reward design based on training behavior, and progressively improve task satisfaction across RL stages.

\subsection{Optimization via Denoising Diffusion Policy Optimization}
In order to optimize the pretrained diffusion model against the synthesized rewards, we execute the \textbf{Diffusion RL Loop}. We treat the diffusion sampling process as a multi-step Markov Decision Process (MDP) and employ Denoising Diffusion Policy Optimization (DDPO) \cite{black2023trainingdiffusionmodelsrl} to update the model weights. 

During the Reward Evaluation phase, the total reward $R_{\text{total}}$ for each generated scene is computed as a weighted sum of the Universal ($R^u$) and Task-Specific ($R^t$) reward terms, with scalar weights $w^u$ and $w^t$ respectively:
\begin{equation}
    R_{\text{total}} = \sum_{R^u} w^u R^u + \sum_{R^t} w^t R^t
\end{equation}
The weights $w^u$ and $w^t$ are chosen by the LLM and adjusted through reward reflection, described next in Sec~\ref{sec:reward_reflection}.
By utilizing RL, we successfully optimize the generator for complex, non-differentiable geometric constraints that standard gradient-based guidance methods cannot handle.

\subsection{Reward Reflection and Iterative Refinement}
\label{sec:reward_reflection}
A key challenge in reward-guided generation is the potential for "reward hacking" or the formulation of objectives that are too sparse for the current model state. iARCS addresses this through the \textbf{Reward Reflection} module. 

After a fixed number of RL iterations, the LLM inspects the \textbf{Reward Statistics} produced by the current reward program together with its \textbf{Evolution Memory}: the full history of reward programs written for this task and the statistics each one produced during training. Top-down projections of the generated scenes are provided alongside these statistics as additional information. From this history the LLM can tell whether a reward is improving, saturated, or stuck, and may either:
\begin{itemize}
    \item \textbf{Redefine:} Debug and simplify the revised reward functions if the underlying logic is flawed or too hard for current policy.
    \item \textbf{Curriculum Generation:} Decompose the reward into a simpler, intermediate objective to ``warm up" the model before introducing the full constraint.
\end{itemize}
This iterative feedback loop ensures the generator progressively learns to satisfy high-level user intent without collapsing the diversity of the scene prior.
The full method is summarized in Algorithm~\ref{alg:ddpo-is}.

\begin{algorithm}[h]
\caption{DDPO Training for 3D Scene Generation}
\label{alg:ddpo-is}
\SetKwInOut{Input}{Input}
\SetKwInOut{Output}{Output}
\small
\Input{Pretrained policy $\pi_\theta$, floor plan $F$, universal reward $R^u$, task reward $R^t$, weights $w^u, w^t$}
\Output{Adapted LoRA policy $\pi_{\theta_C}$}

\For{$\text{RL stage} = 1$ \KwTo $N_{\text{stages}}$}{
    $\text{trajectories} \gets []$\;
    $\mathbf{x}_T \sim \mathcal{N}(\mathbf{0}, \mathbf{I})$ \tcp*{batch of $B$ rollouts}
    
    \For{$t = T$ \KwTo $1$}{
        $\boldsymbol{\epsilon}_{\text{pred}} \gets \text{model}(\mathbf{x}_t, F, t)$\;
        $\mathbf{x}_{t-1}, \log p_{\text{old}} \gets \text{DDIM\_step}(\boldsymbol{\epsilon}_{\text{pred}}, \mathbf{x}_t, t)$\;
        $\text{trajectories.append}((\mathbf{x}_t, \log p_{\text{old}}, t))$\;
        $\mathbf{x}_t \gets \mathbf{x}_{t-1}$\;
    }
    $\mathbf{x}_0 \gets \mathbf{x}_t$ \tcp*{final denoised scene}
    
    $r^u \gets R^u(\mathbf{x}_0, F)$\;
    $r^t \gets R^t(\mathbf{x}_0, F)$\;
    $r_{\text{total}} \gets w^u \cdot r^u + w^t \cdot r^t$\;
    $A \gets \text{normalize}(r_{\text{total}})$\;
    
    \For{$\text{PPO update} = 1$ \KwTo $N_{\text{updates}}$}{
        $\mathcal{B} \gets \text{sample\_timesteps}(\text{trajectories})$ \tcp*{$15$ of $T$}
        \For{$i \in \mathcal{B}$}{
            $\mathbf{x}_t, \log p_{\text{old}}, t \gets \text{trajectories}[i]$\;
            $\boldsymbol{\epsilon}_{\text{new}} \gets \text{model}(\mathbf{x}_t, F, t)$\;
            $\log p_{\text{new}} \gets \text{DDIM\_logprob}(\boldsymbol{\epsilon}_{\text{new}}, \mathbf{x}_t, t)$\;
            $\rho \gets \exp(\log p_{\text{new}} - \log p_{\text{old}})$\;
            $\mathcal{L} \gets -\mathbb{E}[\rho \cdot A]$\;
            Update model parameters via $\nabla_\theta \mathcal{L}$\;
        }
    }
}
\end{algorithm}

\section{Experiments}
\label{sec:experiments}

To demonstrate the viability of iARCS as a scalable synthetic data generator, our experiments focus on three core questions: (1) Can iARCS improve physical and design constraint satisfaction while preserving the original data diversity? (2) Does our generated data improve MiDiffusion trained with only 3D-FRONT? (3) Can the agentic framework handle specific, task-oriented constraints?

\subsection{Experimental Setup}
\label{subsec:setup}

\noindent\textbf{Dataset and Base Model.} We use 3D-FRONT~\cite{3d_front}, a synthetic dataset of 6,813 professionally designed indoor scenes, with the standard split from prior work~\cite{midiffusion,atiss,diffuscene}. Our base generator is a continuous domain-only MiDiffusion model~\cite{midiffusion} pretrained on 3D-FRONT; it operates in object-parameter space and retrieves canonical CAD models from 3D-FUTURE~\cite{3d_future}. Unless otherwise stated, experiments use 3D-FRONT bedroom scenes.

\noindent\textbf{RL Finetuning.} We fine-tune with LoRA~\cite{hu2021loralowrankadaptationlarge} ($r=16$, $\alpha=16$), 20-step DDIM rollouts~\cite{song2020ddim} with $\eta=1.0$, and DDPO with importance sampling~\cite{black2023trainingdiffusionmodelsrl}. We use Adam~\cite{kingma2017adammethodstochasticoptimization} with learning rate $3\times10^{-4}$; full-model fine-tuning of this generator needs $10^{-5}$, but with LoRA that rate reaches the same result far more slowly. Rather than stopping at a fixed step, we track reward jointly with the plausibility and functional metrics and select an operating point on the resulting trade-off curve; the criterion is given in Appendix~\ref{app:rl}.

\noindent\textbf{Rewards.} The first is a set of universally applicable rewards that improve overall physical plausibility and functional utility. These rewards are manually designed targeting collision avoidance and boundary adherence; reachability and walkability are never optimized and so remain held-out measurements. Full implementation details are given in Appendix~\ref{app:rl}. The second type is task-specific rewards generated from user prompts by our agentic pipeline; we use Gemini 3.6 Flash as the reward-synthesis backend, and report a sensitivity study with GPT-5.4 in Appendix~\ref{app:rl} (Tab.~\ref{tab:llm_backend_supp}). Appendix~\ref{app:agentic} describes the prompts used to generate executable reward functions.

\noindent\textbf{Baselines.} We compare against ATISS~\cite{atiss} (autoregressive), MiDiffusion~\cite{midiffusion} (continuous-only mixed diffusion), InstructScene~\cite{lin2024instructsceneinstructiondriven3dindoor} (LLM/graph-driven, not floor-plan conditioned) and PhyScene~\cite{physcene} (differentiable guidance). No PhyScene training code is released, so we reimplemented and retrained it on our splits. All methods are scored with the same harness on identical splits. We do not retrain Holodeck~\cite{yang2024holodecklanguageguidedgeneration} or Steerable Scene Generation~\cite{pfaff2025steerablescenegenerationpost}, since they cannot be trained on the same footing: Holodeck places geometry with an LLM, and Steerable Scene Generation searches at inference time (Sec.~\ref{sec:related}).

\noindent\textbf{Evaluation Metrics.}
\label{subsec:metrics}
Following prior scene-generation work~\cite{atiss,diffuscene,midiffusion,physcene}, we report distribution metrics on 1,080 synthesized scenes: CLIP-FID$\downarrow$ and Scene Classification Accuracy (SCA$\downarrow$, best near the 50\% chance level). We use CLIP features rather than InceptionNet because our inputs are top-down projections. Physical plausibility and functional utility are measured directly in 3D: object- and scene-collision rates $Col_{obj}\downarrow$, $Col_{scene}\downarrow$ (any non-zero bounding-box overlap counts as a collision), out-of-bound rate $R_{out}\downarrow$, reachable-object ratio $R_{reach}\uparrow$, walkability $R_{walkable}\uparrow$ (largest connected walkable region over total walkable area), and Success$\uparrow$, the fraction of scenes satisfying the task constraint under a binary geometric check applied identically to every method. Full definitions are in Appendix~\ref{app:metrics}.

We evaluate task adherence and physical realism using 3D-FRONT scenes filtered by each task constraint.

\subsection{Scene Synthesis}

Tab.~\ref{tab:main_comparison} reports quantitative results: iARCS is best among the generated methods on every metric, spanning physical plausibility, functional utility and distribution-level quality. These gains are not from placing fewer objects: across the 4{,}000-scene release set iARCS averages 4.92, 9.10 and 13.66 objects per bedroom, living and dining scene, in the range of the 5.15, 12.01 and 11.12 in the data. Rejection sampling from the pretrained model, the cheapest training-free alternative, draws $3.7\times$ more scenes yet is worse than iARCS on CLIP-FID and SCA in every room (Appendix~\ref{app:rejection}).

\begin{figure}[h]
\centering
\includegraphics[width=\columnwidth]{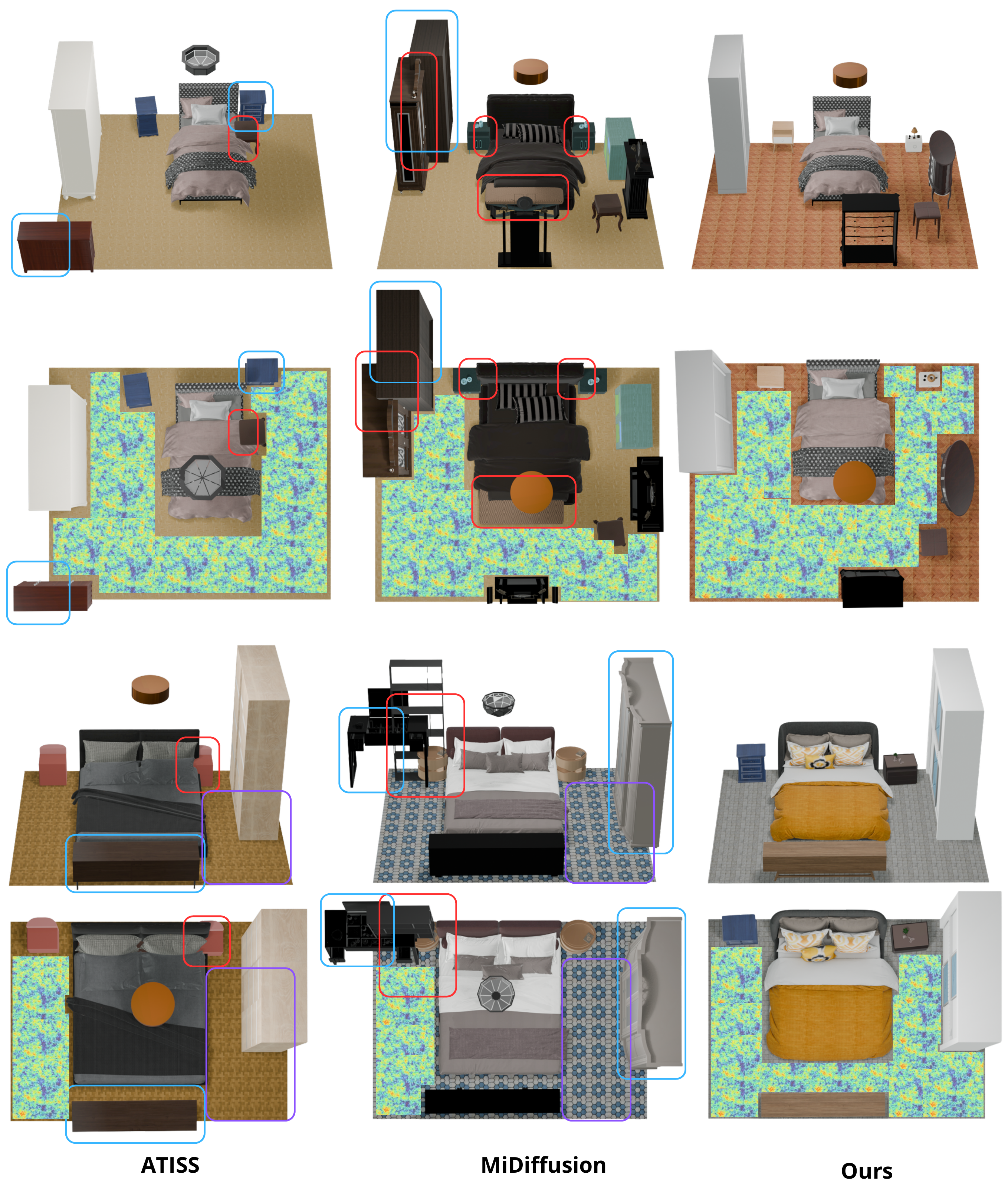}
\caption{\textbf{Qualitative scene synthesis comparison.} Compared with ATISS and MiDiffusion, iARCS produces layouts with fewer collisions, better boundary adherence, and clearer walkable regions; red, blue, and purple boxes mark object collisions, out-of-bound objects, and unreachable areas.}
\label{fig:qualitative_main}
\end{figure}

Qualitative comparisons in Fig.~\ref{fig:qualitative_main} show that iARCS produces layouts with fewer collisions, better boundary adherence, and improved walkable free space than ATISS and MiDiffusion.

\begin{figure}[h]
\centering
\includegraphics[width=\columnwidth]{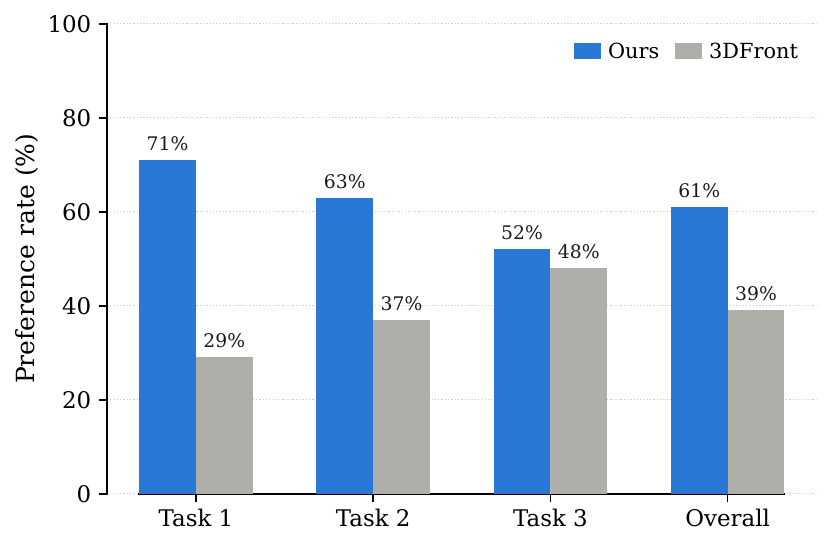}
\caption{\textbf{Human preference comparison.} Preference rates for iARCS versus constraint-filtered 3D-FRONT scenes, using the same Tasks 1--3 as Table~\ref{tab:taskwise_effectiveness}.}
\label{fig:preference_rate}
\end{figure}

\noindent\textbf{Human Evaluation.} For each task, we sampled 20 constraint-filtered 3D-FRONT scenes and 20 iARCS scenes. Each of 21 users selected the 10 best scenes for each constraint by task satisfaction and scene plausibility; Fig.~\ref{fig:preference_rate} reports preference rates.

\begin{table}[t]
\centering
\caption{\textbf{Average scene-synthesis performance.} Each metric is averaged across the bedroom, living-room, and dining-room evaluations; per-room results are in Appendix~\ref{app:roomwise}. All values are percentages except $R_{walkable}$ and CLIP-FID.}
\label{tab:main_comparison}
\scriptsize
\setlength{\tabcolsep}{2pt}
\renewcommand{\arraystretch}{1.1}
\resizebox{\columnwidth}{!}{%
\begin{tabular}{lccccccc}
\toprule
Method & $Col_{obj}\downarrow$ & $Col_{scene}\downarrow$ & $R_{out}\downarrow$ & $R_{reach}\uparrow$ & $R_{walkable}\uparrow$ & CLIP-FID$\downarrow$ & SCA$\downarrow$ \\
\midrule
Ground Truth (GT) & 43.12 & 80.08 & 2.67 & 71.51 & 0.851 & -- & -- \\
ATISS & 71.77 & 82.35 & 19.86 & 65.24 & 0.814 & 2.46 & 82.83 \\
MiDiffusion & 62.98 & 91.49 & 17.45 & 77.35 & 0.829 & 2.50 & 77.29 \\
InstructScene & 52.95 & 88.04 & 43.39 & 69.22 & 0.843 & 5.60 & 83.97 \\
PhyScene & 53.87 & 88.19 & 24.29 & 70.55 & 0.849 & 2.48 & 78.62 \\
\textbf{iARCS (Ours)} & \textbf{50.04} & \textbf{74.88} & \textbf{10.99} & \textbf{80.39} & \textbf{0.896} & \textbf{2.45} & \textbf{76.49} \\
\bottomrule
\end{tabular}%
}
\end{table}

\subsection{Data Augmentation Improves Base Generator}
\label{subsec:augmentation}
Table~\ref{tab:generated_data_comparison} shows that augmenting 3D-FRONT with 4,000 iARCS-generated scenes improves every plausibility and functional metric for both ATISS and MiDiffusion, at essentially unchanged CLIP-FID and SCA. Augmenting with the same number of unsteered MiDiffusion scenes helps less, so the gain comes from the constraint-steered scenes rather than from additional synthetic data alone. We generate scenes on training-set floor layouts $F$ and fine-tune all weights for 200 epochs at learning rate $10^{-5}$. Fig.~\ref{fig:augmentation_qualitative} shows improved feasibility and accessibility; dataset-release details are in Appendix~\ref{app:release}.

\begin{table}[h]
\centering
\caption{\textbf{Augmentation with iARCS versus baseline-generated data.} ATISS and MiDiffusion trained on 3D-FRONT, on 3D-FRONT plus 4{,}000 unsteered MiDiffusion scenes, and on 3D-FRONT plus 4{,}000 iARCS scenes. iARCS augmentation improves every plausibility and functional metric over the unaugmented model at essentially unchanged diversity, and gives the larger gain throughout.}
\label{tab:generated_data_comparison}
\setlength{\tabcolsep}{4pt}
\resizebox{\columnwidth}{!}{%
\renewcommand{\arraystretch}{1.25}
\begin{tabular}{l|ccc|cc|cc}
\toprule
\multirow{2}{*}{Method} & \multicolumn{3}{c|}{Physical Plausibility} & \multicolumn{2}{c|}{Functional Utility} & \multicolumn{2}{c}{Diversity} \\
\cline{2-8}
& $Col_{obj}\downarrow$ & $Col_{scene}\downarrow$ & $R_{out}\downarrow$ & $R_{reach}(\%)\uparrow$ & $R_{walkable}\uparrow$ & CLIP-FID$\downarrow$ & SCA$\downarrow$ \\
\midrule
ATISS (3D-FRONT) & 72.39\% & 53.61\% & 16.35\% & 60.27\% & 0.839 & \textbf{1.52} & 79.24\% \\
ATISS (3D-FRONT + MiDiffusion aug.) & 71.54\% & 51.39\% & 17.82\% & 60.92\% & 0.839 & 1.54 & 78.82\% \\
ATISS (3D-FRONT + iARCS aug.) & \textbf{69.60\%} & \textbf{50.28\%} & \textbf{13.21\%} & \textbf{64.59\%} & \textbf{0.862} & \textbf{1.52} & \textbf{78.07\%} \\
\midrule
MiDiffusion (3D-FRONT) & 52.67\% & 81.67\% & 5.89\% & 85.70\% & 0.806 & \textbf{1.34} & \textbf{65.99\%} \\
MiDiffusion (3D-FRONT + MiDiffusion aug.) & 48.26\% & 75.65\% & 3.78\% & 90.00\% & 0.813 & 1.37 & 66.86\% \\
MiDiffusion (3D-FRONT + iARCS aug.) & \textbf{41.49\%} & \textbf{63.61\%} & \textbf{3.12\%} & \textbf{92.52\%} & \textbf{0.827} & \textbf{1.34} & 66.10\% \\
\bottomrule
\end{tabular}%
}
\end{table}

\begin{figure}[h]
\vspace{-1em}
\centering
\includegraphics[width=\columnwidth]{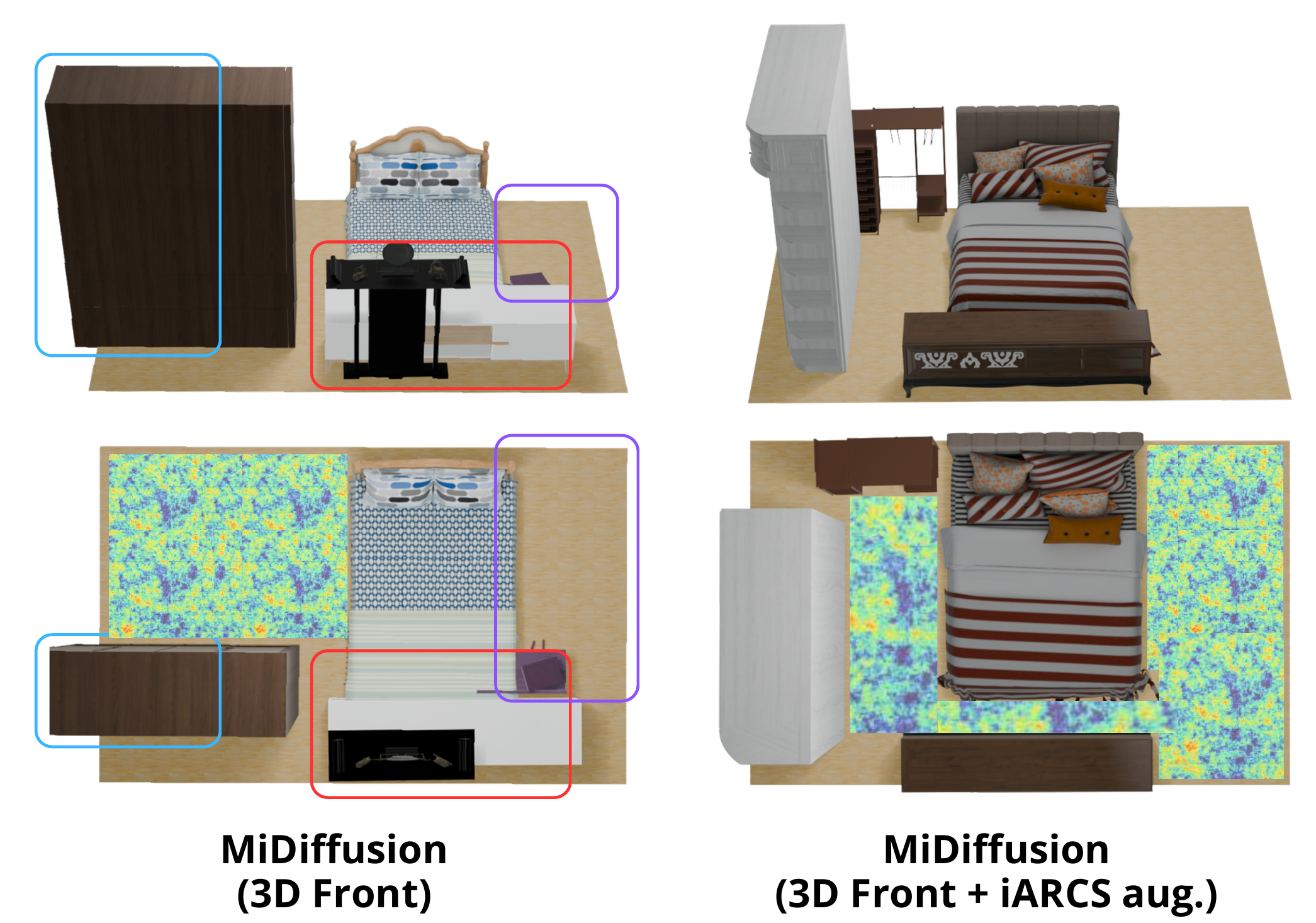}
\caption{\textbf{Effect of iARCS data augmentation.} MiDiffusion trained with 3D-FRONT + iARCS-generated data improves layout feasibility and accessibility while preserving realism; red, blue, and purple boxes mark collisions, out-of-bound objects, and unreachable areas.}
\label{fig:augmentation_qualitative}
\vspace{-1em}
\end{figure}

\subsection{Task-Specific Constraint Optimization}

Tab.~\ref{tab:taskwise_effectiveness} reports quantitative results on complex, task-specific optimization settings. Here, 3D-FRONT* denotes the subset of 3D-FRONT scenes that satisfy each task constraint, and MiDiffusion is the unsteered base model filtered to the same number of satisfying scenes; Success is measured over the full pool, all other metrics on these matched sets. iARCS satisfies task constraints far more often than the baselines, reduces collisions relative to the base model, and maintains diversity comparable to 3D-FRONT*. This demonstrates its utility for constraint-aware synthetic data generation and augmentation.

iARCS uses the LLM to interpret complex tasks, decompose them into optimizable constraints and synthesize executable rewards; Appendix~\ref{app:reward_example} gives a qualitative example of the resulting layouts.

\begin{table}[t]
\centering
\caption{\textbf{Task-wise constrained generation results.} Task~1: ``robot grasping scene where all support surfaces are within a 1.0\,m vertical reach limit''; Task~2: ``room with a TV stand positioned so a farsighted person can view a 4K TV from bed''; Task~3: ``a bedroom scene with a functional study zone''. 3D-FRONT$^{*}$ (grey) is the constraint-satisfying subset of real scenes, shown for reference rather than as a competing method. Success is measured over the full generation pool, all other metrics on the matched satisfying sets. Bold marks where iARCS improves on the base model it adapts, which holds for every metric in every task.}
\label{tab:taskwise_effectiveness}
\setlength{\tabcolsep}{2pt}
\footnotesize
\resizebox{\columnwidth}{!}{%
\renewcommand{\arraystretch}{1.05}
\begin{tabular}{cl|c|c|ccc|cc}
\toprule
\multirow{2}{*}{Task} & \multirow{2}{*}{Method} & \multirow{2}{*}{Success} & Diversity & \multicolumn{3}{c|}{Physical Plausibility} & \multicolumn{2}{c}{Functional Utility} \\
\cline{4-9}
& & & SCA$\downarrow$ & Col$_{obj}\downarrow$ & Col$_{scene}\downarrow$ & $R_{out}\downarrow$ & $R_{reach}\uparrow$ & $R_{walkable}\uparrow$ \\
\midrule
\multirow{3}{*}{1} & 3D-FRONT$^{*}$ &  \textcolor{gray}{5.67\%} & \textcolor{gray}{97.34\%} & \textcolor{gray}{50.68\%} & \textcolor{gray}{83.72\%} & \textcolor{gray}{2.79\%} & \textcolor{gray}{82.26\%} & \textcolor{gray}{0.786} \\
 & MiDiffusion &  4.53\% & 86.30\% & 54.29\% & 88.80\% & 6.19\% & 75.40\% & 0.711 \\
 & iARCS (Ours) &  \textbf{73.83\%} & \textbf{77.92\%} & \textbf{34.77\%} & \textbf{59.28\%} & \textbf{5.84\%} & \textbf{80.37\%} & \textbf{0.744} \\
\midrule
\multirow{3}{*}{2} & 3D-FRONT$^{*}$ &  \textcolor{gray}{6.09\%} & \textcolor{gray}{97.02\%} & \textcolor{gray}{38.79\%} & \textcolor{gray}{70.04\%} & \textcolor{gray}{1.71\%} & \textcolor{gray}{86.38\%} & \textcolor{gray}{0.810} \\
 & MiDiffusion &  5.22\% & 89.34\% & 56.10\% & 78.98\% & 5.78\% & 79.32\% & 0.733 \\
 & iARCS (Ours) &  \textbf{66.02\%} & \textbf{85.70\%} & \textbf{46.46\%} & \textbf{77.15\%} & \textbf{3.45\%} & \textbf{86.77\%} & \textbf{0.790} \\
\midrule
\multirow{3}{*}{3} & 3D-FRONT$^{*}$ &  \textcolor{gray}{1.04\%} & \textcolor{gray}{97.58\%} & \textcolor{gray}{56.86\%} & \textcolor{gray}{88.67\%} & \textcolor{gray}{3.33\%} & \textcolor{gray}{75.17\%} & \textcolor{gray}{0.728} \\
 & MiDiffusion &  0.87\% & 85.40\% & 72.00\% & 98.52\% & 5.10\% & 69.32\% & 0.704 \\
 & iARCS (Ours) &  \textbf{33.39\%} & \textbf{77.97\%} & \textbf{52.38\%} & \textbf{79.39\%} & \textbf{4.59\%} & \textbf{74.75\%} & \textbf{0.738} \\
\bottomrule
\end{tabular}%
}
\end{table}

\subsection{Ablations}
\begin{figure}[h]
\centering
\includegraphics[width=\columnwidth]{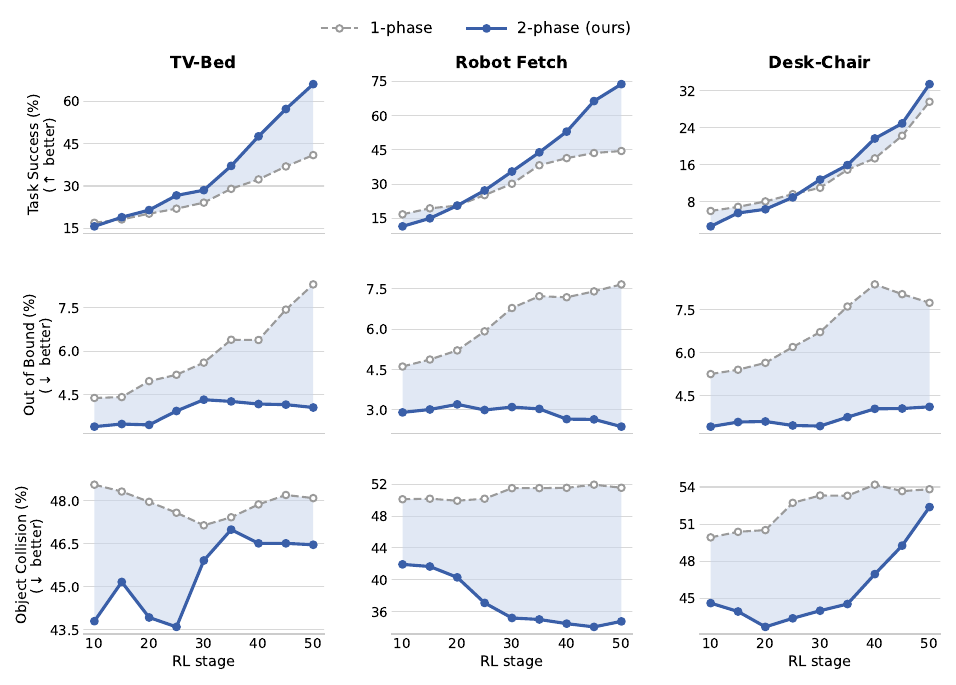}
\caption{\textbf{Single-phase vs. two-phase training.} Two-phase training first optimizes physical and functional quality, then task constraints, yielding more feasible and usable layouts.}
\label{fig:ablation_2stage}
\end{figure}

Fig.~\ref{fig:ablation_2stage} ablates two-phase training under the same reward level. Directly optimizing task, physical, and functional rewards performs worse than first training with universal rewards, then jointly fine-tuning with task rewards.

\noindent\textbf{Isolating the agentic loop.} Tab.~\ref{tab:mechanism_ablation} separates the quality of the reward from the process that produces it, over three tasks and three seeds at matched base model and budget. Freezing the initial reward for the whole run is weakest, reaching 28.4 on Task~1. Freezing the final reward from step~0 reaches 69.1, close to the full method at 73.83, but this reward is itself an output of the loop and is not available in advance. Removing the evolution memory while retaining iteration reaches 35.3, so iteration on its own recovers little. Appendix~\ref{app:iterative} shows the same effect across all three tasks: single-pass rewards are often overly complex or poorly shaped, while refinement inspects failures and revises the reward, improving task adherence without loss of plausibility.

\begin{table}[h]
\centering
\caption{\textbf{Isolating the agentic mechanism.} Task success rate (\%, $\uparrow$), mean$\pm$std over 3 seeds. Freezing the final reward at step~0 tests reward quality alone; the no-memory variant keeps iteration but discards previous reward programs and their statistics.}
\label{tab:mechanism_ablation}
\footnotesize
\setlength{\tabcolsep}{3pt}
\resizebox{\columnwidth}{!}{%
\begin{tabular}{lccc}
\toprule
Method & Task 1 & Task 2 & Task 3 \\
\midrule
3D-FRONT$^{*}$ & 5.67 & 6.09 & 1.04 \\
MiDiffusion & 4.53 & 5.22 & 0.87 \\
\midrule
Initial reward, frozen & $28.4 \pm 5.1$ & $22.7 \pm 4.3$ & $9.6 \pm 2.7$ \\
Final reward, frozen at step 0 & $69.1 \pm 4.6$ & $61.8 \pm 5.2$ & $32.2 \pm 4.1$ \\
Iterative, no memory & $35.3 \pm 6.8$ & $41.4 \pm 7.1$ & $12.8 \pm 5.6$ \\
iARCS (full) & $\mathbf{73.83 \pm 3.2}$ & $\mathbf{66.02 \pm 3.8}$ & $\mathbf{33.39 \pm 3.5}$ \\
\bottomrule
\end{tabular}%
}
\end{table}

\subsection{Limitations}
\label{subsec:limitations}
Reward quality depends on the LLM's task decomposition and code synthesis, so ambiguous prompts can yield incomplete constraints. Two-phase training adds compute cost from iterative RL and repeated reward evaluation. LoRA adapters do not transfer: each constraint needs its own run and adapter. Finally, we evaluate only on 3D-FRONT-style indoor layouts and do not demonstrate compositional or sequential constraints.

\section{Conclusion}
We presented iARCS, an iterative agentic RL framework for controllable 3D indoor scene generation that combines LLM-based task decomposition, executable reward synthesis, and phased policy refinement. Experiments show that iARCS improves physical plausibility and functional utility while maintaining competitive distribution quality. Moreover, iARCS-generated data can further improve a base scene generator through self-augmentation. We also demonstrated strong task-conditioned generation, where iARCS achieves lower SCA than constraint-satisfying dataset subsets, indicating improved diversity under user-specified constraints.

For future work, we aim to extend generalization beyond 3D-FRONT-style indoor distributions to broader domains, including physically grounded 4D generation tasks.

\clearpage
\newpage
{
    \small
    \bibliographystyle{ieeenat_fullname}
    \bibliography{main}
}

\clearpage
\appendix
\section*{Appendix}
\addcontentsline{toc}{section}{Appendix}
\section{Base Diffusion Model}
\label{app:base}
\noindent\textbf{Object and Floor Plan Encoding.} Following MiDiffusion~\cite{midiffusion}, we encode object features by processing geometric attributes through an MLP and combining them with learnable class label embeddings. For floor plan conditioning, we sample 256 boundary points from each floor plan image and compute their outward-facing normal vectors. These floor plan features are extracted using a PointNet-based~\cite{qi2017pointnetdeeplearningpoint} encoder adapted from LEGO-Net~\cite{wei2023legonetlearningregularrearrangements}.

\noindent\textbf{Denoising Network Architecture.} The denoising network uses a Transformer architecture with 8 layers, an embedding dimension of 512, 4 attention heads, a feed-forward dimension of 2048, and a dropout rate of 0.1. We use GELU~\cite{hendrycks2023gaussianerrorlinearunits} activation, adaptive layer normalization with absolute timestep encoding, and fully-connected MLP layers. The feature extractor is a PointNet-based architecture with layer dimensions [4, 64, 64, 512, 64].

\section{RL Finetuning}
\label{app:rl}

\noindent\textbf{Denoising Diffusion Policy Optimization (DDPO).} To align the generative process with complex, non-differentiable spatial constraints, we employ DDPO \cite{black2023trainingdiffusionmodelsrl}. This framework treats the iterative denoising process as a multi-step Markov Decision Process (MDP) where the denoising network acts as the policy. 

\noindent\textbf{LoRA Configuration.} To facilitate efficient fine-tuning, we integrate Low-Rank Adaptation (LoRA) \cite{hu2021loralowrankadaptationlarge} into all attention projection layers (\texttt{q\_proj}, \texttt{k\_proj}, \texttt{v\_proj}, \texttt{out\_proj}) within the Transformer blocks. We use a rank $r=16$ and scaling factor $\alpha=16$. This setup focuses the optimization on a small subset of parameters, preventing catastrophic forgetting of the base model's structural priors while providing enough capacity to learn the target reward surfaces.

\noindent\textbf{Training Configuration.} We use the Adam optimizer \cite{kingma2017adammethodstochasticoptimization} with a learning rate of $3 \times 10^{-4}$ and no weight decay, and a batch size of 32 for both the sampling (rollout) and training phases. Full-model fine-tuning of this generator requires a learning rate of $10^{-5}$; with LoRA, $10^{-5}$ reaches the same results but converges considerably more slowly, so we use $3 \times 10^{-4}$ throughout. All experiments run on a single NVIDIA GeForce RTX 2080 Ti GPU with mixed-precision FP16.

\noindent\textbf{Policy Optimization.} We use \emph{phase} for the two optimization phases and \emph{RL stage} for one rollout-generation and policy-update cycle. Each RL stage samples a batch of 32 rollouts, each a 20-step denoising trajectory, and performs a single PPO update on 15 timesteps drawn uniformly at random from the 20 steps of each trajectory. Reward reflection is performed every 10 RL stages in Phase~2; Phase~1 uses fixed universal rewards and no reflection. For bedrooms the denoising state has shape $(B, N, 30)$: eight geometric dimensions and 22 category logits, and the DDIM log-probability is taken over all thirty, so categories are optimized alongside geometry.

\noindent\textbf{Stopping Criteria.} Phase~1 runs for a fixed budget of 100 RL stages. We then plot the trade-off (Pareto) curve over the tracked metrics and select an operating-point checkpoint rather than the last one, since further reward gain comes at the cost of other metrics. Throughout both phases we track mean reward together with mean collision rate, mean out-of-boundary rate, and the remaining functional metrics, and assign a tolerance band to each; a run is terminated if reward continues to rise while any tracked metric degrades beyond its tolerance, which is our operational test for reward hacking. Phase~2 uses the same monitor-and-select procedure on top of the task reward.

\noindent\textbf{Reward-Code Reliability and Inference Cost.} Across all experiments reported here, every reward program written by the LLM executed without error, so no retry or repair round was needed. The agentic loop runs only at training time: inference is a standard diffusion pass with no LLM calls. Generation is a 20-step denoising pass, batched at 32 scenes, and costs $\approx$1.2\,ms per scene on a single RTX 2080 Ti under FP16; end-to-end generation including scene serialization produces on the order of 1{,}000 scenes in a few seconds. Applying a LoRA adapter adds no measurable overhead over the base generator.

\noindent\textbf{Compute and Token Cost.} Tab.~\ref{tab:compute_cost} reports the wall-clock time and LLM usage per task on a single RTX 2080 Ti, with 0.57\,GB peak GPU memory. Reward reflection adds about 15 minutes and five LLM calls per task over training with a fixed reward.

\begin{table}[h]
\centering
\caption{\textbf{Compute and token cost per task.} Mean$\pm$std over all runs.}
\label{tab:compute_cost}
\small
\setlength{\tabcolsep}{3pt}
\resizebox{\columnwidth}{!}{%
\begin{tabular}{lccccc}
\toprule
Reward source & RL training & Reflection & Total & LLM calls & Tokens \\
\midrule
Fixed reward (no LLM) & $27.2 \pm 1.4$ min & -- & $27.2 \pm 1.4$ min & 0 & 0 \\
Agentic (GPT-5.4) & $26.1 \pm 1.9$ min & $17.2 \pm 1.6$ min & $43.5 \pm 4.5$ min & 5 & 160k $\pm$ 14k \\
Agentic (Gemini 3.6 Flash) & $27.1 \pm 0.8$ min & $14.12 \pm 1.6$ min & $41.2 \pm 1.8$ min & 5 & 168k $\pm$ 12k \\
\bottomrule
\end{tabular}%
}
\end{table}

\noindent\textbf{Reward Modeling.} We adopt a hybrid reward strategy. A Gemini 3.6 Flash model serves as the reward-synthesis backend, writing and revising executable reward programs from the task prompt and reward statistics; it never inspects generated scenes directly. This is augmented by a set of "Universal Rewards" which provide dense, objective signals. These rewards, detailed in Algorithms~\ref{alg:collision}--\ref{alg:count}, explicitly penalize physical inconsistencies such as object collisions and boundary violations while encouraging realistic object density and accessibility. For iterative agentic reward improvement, we perform reward reflection and reward-code refinement every 10 RL stages.

\noindent\textbf{Sensitivity to the Reward-Synthesis Backend.} Because every task reward is written by an LLM, we repeat reward synthesis with a second backend under the same base model, training budget, and task prompts (Tab.~\ref{tab:llm_backend_supp}). Both backends produce runnable reward programs and reach comparable success on all three tasks, so the method is not specific to one proprietary model; Gemini 3.6 Flash is somewhat stronger and is used for the main results.

\begin{table}[h]
\centering
\caption{\textbf{Sensitivity to the LLM reward-synthesis backend.} Task success rate (\%, $\uparrow$), mean$\pm$std over 3 random seeds.}
\label{tab:llm_backend_supp}
\small
\setlength{\tabcolsep}{4pt}
\begin{tabular}{lccc}
\toprule
LLM backend & Task 1 & Task 2 & Task 3 \\
\midrule
GPT-5.4 & $65.13 \pm 1.2$ & $63.3 \pm 3.69$ & $31.18 \pm 2.48$ \\
Gemini 3.6 Flash & $\mathbf{73.83 \pm 3.2}$ & $\mathbf{66.02 \pm 3.8}$ & $\mathbf{33.39 \pm 3.5}$ \\
\bottomrule
\end{tabular}
\end{table}

\begin{algorithm}[h]
\small
\caption{Collision Avoidance Reward}
\label{alg:collision}
\KwIn{Generated scene $\mathbf{x}_0$}
\KwOut{Collision reward $r_{\text{collision}}$}

positions, sizes, classes $\gets$ parse\_scene($\mathbf{x}_0$)\;
ground\_objects $\gets$ filter\_ceiling(classes)\;
total\_penetration $\gets 0$\;

\For{each pair $(i, j)$ in ground\_objects}{
    center\_dist $\gets |$positions$[i] -$ positions$[j]|$\;
    sum\_extents $\gets$ sizes$[i] +$ sizes$[j]$\;
    overlap\_per\_axis $\gets$ sum\_extents $-$ center\_dist\;
    penetration $\gets \min($overlap\_per\_axis$)$\;
    \If{penetration $> 0$}{
        total\_penetration $\gets$ total\_penetration $+$ penetration\;
    }
}
total\_penetration $\gets$ total\_penetration $/$ 2\;
\Return{$+1$ if total\_penetration $= 0$ else $-$total\_penetration}\;
\end{algorithm}

\begin{algorithm}[h]
\small
\caption{Boundary Adherence Reward}
\label{alg:boundary}
\KwIn{Scene $\mathbf{x}_0$, floor polygon}
\KwOut{Boundary reward $r_{\text{boundary}}$}

positions, sizes, orientations $\gets$ parse\_scene($\mathbf{x}_0$)\;
total\_violation $\gets 0$\;

\For{each object $i$}{
    corners $\gets$ compute\_oriented\_bbox(positions$[i]$, sizes$[i]$, orientations$[i]$)\;
    object\_polygon $\gets$ Polygon(corners)\;
    intersection\_area $\gets$ area(object\_polygon $\cap$ floor\_polygon)\;
    object\_area $\gets$ area(object\_polygon)\;
    oob\_area $\gets$ object\_area $-$ intersection\_area\;
    total\_violation $\gets$ total\_violation $+$ oob\_area\;
}
\Return{$+2$ if total\_violation $= 0$ else $-$total\_violation}\;
\end{algorithm}

\begin{algorithm}[h]
\small
\caption{Accessibility Reward}
\label{alg:accessibility}
\KwIn{Scene $\mathbf{x}_0$, floor grid}
\KwOut{Accessibility reward $r_{\text{access}}$}

occupancy\_grid $\gets$ floor\_grid.copy()\;
positions, sizes $\gets$ parse\_scene($\mathbf{x}_0$)\;

\For{each object $i$ (excluding ceiling)}{
    footprint $\gets$ get\_footprint(positions$[i]$, sizes$[i]$)\;
    inflated\_footprint $\gets$ inflate(footprint, agent\_radius$=0.3$)\;
    occupancy\_grid[inflated\_footprint] $\gets$ OCCUPIED\;
}

all\_regions $\gets$ find\_connected\_components(occupancy\_grid)\;
largest\_region $\gets \max($all\_regions, key$=$size$)$\;
reachable\_cells $\gets$ flood\_fill(occupancy\_grid, start$=$largest\_region)\;

coverage $\gets$ len(reachable\_cells) $/$ total\_free\_cells\;
num\_regions $\gets$ len(all\_regions)\;
avg\_clearance $\gets$ mean\_distance\_to\_obstacles(reachable\_cells)\;

\Return{$1.0 \cdot$ coverage $- 0.1 \cdot$ num\_regions $+ 0.5 \cdot$ avg\_clearance}\;
\end{algorithm}

\begin{algorithm}[h]
\small
\caption{Object Count Diversity Reward}
\label{alg:count}
\KwIn{Batch of scenes $\{\text{scene}_1, \ldots, \text{scene}_B\}$}
\KwOut{Count diversity reward $r_{\text{count}}$}

object\_counts $\gets$ [count\_objects(scene) for scene in batch]\;
batch\_histogram $\gets$ compute\_histogram(object\_counts, bins$=0$ to $12$)\;
batch\_dist $\gets$ batch\_histogram $/$ batch\_size\;
train\_dist $\gets$ [$0.14$, $0.25$, $0.31$, $0.16$, $0.09$, ...]\;
kl\_divergence $\gets \sum_c$ batch\_dist$[c] \cdot \log($batch\_dist$[c]$ / train\_dist$[c])$\;
\Return{$-$kl\_divergence}\;
\end{algorithm}

\noindent\textbf{Reward Weights.} In our experiments, we use uniform reward weights for universal rewards $R_u$ and task-specific rewards $R_t$, i.e., $R_u = R_t = 1$.

\section{iARCS Dataset}
\label{app:dataset}

\begin{figure*}[h]
\centering
\includegraphics[width=\textwidth]{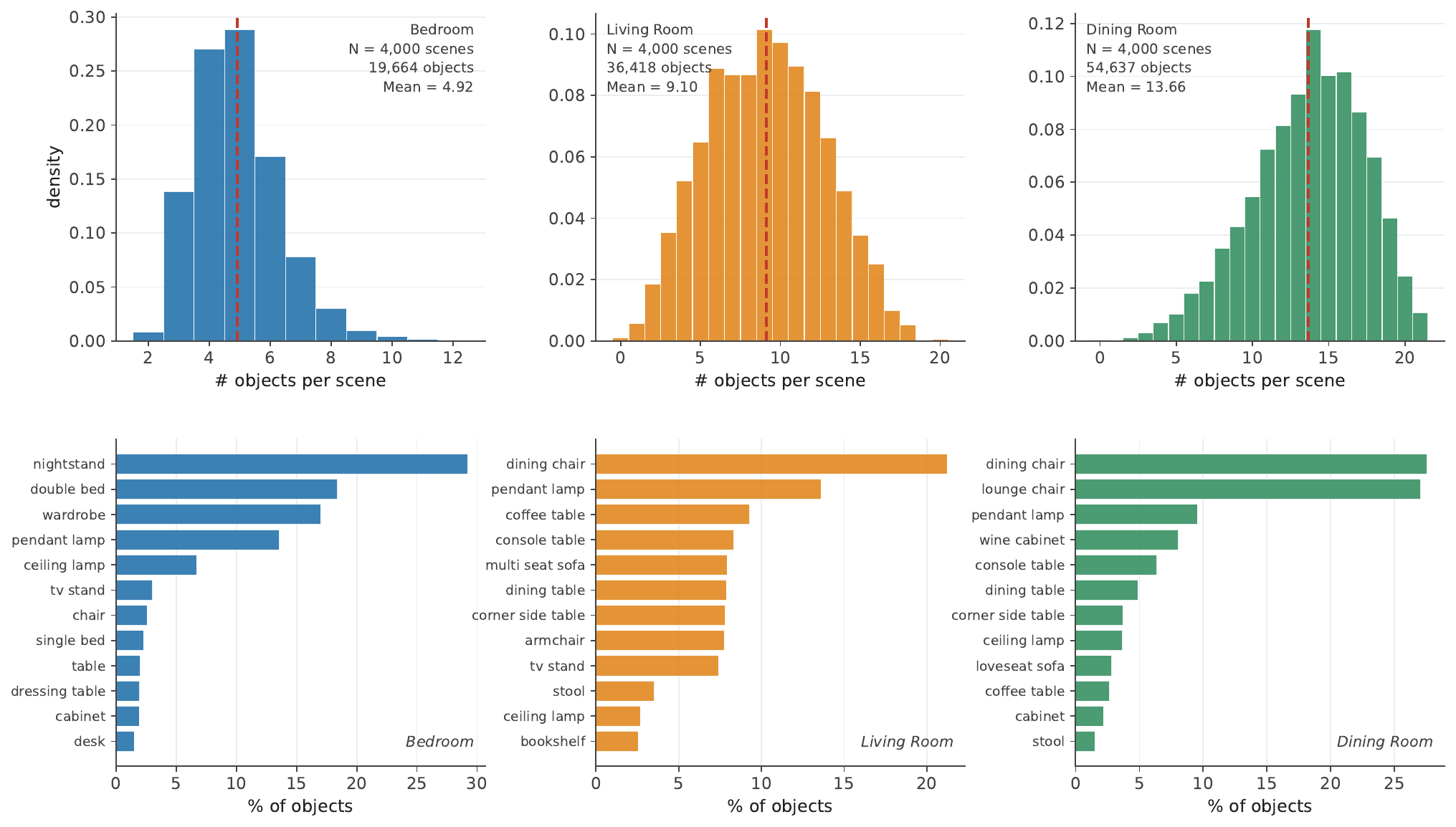}
\caption{\textbf{Distribution of generated scenes across room types.} We sample 4,000 scenes per room. Top: density-normalized histogram of objects per scene (dashed line = mean: Bedroom 4.92, Living 9.10, Dining 13.66). Bottom: the 12 most frequent object categories per room as a percentage of all objects. Counts are unimodal and category frequencies reflect realistic room compositions.}
\label{fig:iarcs_dataset_distribution}
\end{figure*}

We construct an iARCS-generated dataset by refining 4,000 bedroom scenes, 4,000 living-room scenes, and 4,000 dining-room scenes with our method. This dataset provides constraint-refined synthetic layouts for downstream training and evaluation across common indoor room types. Figure~\ref{fig:iarcs_dataset_distribution} shows the distribution of object counts and the most frequent object categories for each room type in the generated dataset.

\subsection{Dataset Release}
\label{app:release}

\noindent\textbf{Contents.} The release contains 12,000 generated scenes: 4,000 bedrooms, 4,000 living rooms and 4,000 dining rooms, produced with the Phase~1 universal-reward model on 3D-FRONT test-split floor plans (162 bedroom, 192 living-room and 177 dining-room layouts). They are separate from the training-set scenes used for augmentation in Sec.~\ref{subsec:augmentation}.

\noindent\textbf{Format.} Scenes are distributed as one record per scene. Each record stores the floor-plan boundary used for conditioning and, for every object, its centroid translation $t_i \in \mathbb{R}^3$, bounding-box dimensions $s_i \in \mathbb{R}^3$, orientation as $(\cos\theta_i, \sin\theta_i)$, and category label, together with the 3D-FUTURE~\cite{3d_future} model identifier used for CAD retrieval. Geometry is therefore reconstructed by retrieval, exactly as in 3D-FRONT~\cite{3d_front}; we redistribute no CAD assets. We will also release the evaluation harness so that all metrics in this paper can be recomputed on the released scenes.

\noindent\textbf{Access and license.} The dataset is hosted on Hugging Face Datasets at \url{https://huggingface.co/datasets/Saugat20021/iARCS} and released under \textbf{CC BY-NC 4.0}. Because the layouts are generated by a model trained on 3D-FRONT and reference 3D-FUTURE assets, the release inherits the non-commercial research terms of those datasets, and users must obtain 3D-FRONT and 3D-FUTURE separately under their own licenses to reconstruct scene geometry.

\noindent\textbf{Intended use and limitations.} The dataset is intended as training and evaluation data for indoor scene synthesis and embodied-AI layout tasks, and specifically as augmentation data for scene generators, the use demonstrated in Sec.~\ref{subsec:augmentation}. It is synthetic throughout and contains no personal, human, or privacy-sensitive content. It inherits the distributional coverage of 3D-FRONT: three residential room types, Western-style furnishing conventions, and the 3D-FUTURE object vocabulary. It should not be treated as a sample of real dwellings, and models trained on it will inherit those biases.

\section{Rejection Sampling Comparison}
\label{app:rejection}

We compare iARCS against the cheapest training-free alternative: sampling more scenes from the pretrained model and keeping the highest-scoring ones under our reward. Table~\ref{tab:rejection} reports the comparison. Despite the recurring sampling cost it is worse on CLIP-FID and SCA in every room, and its lower collision rate comes largely from returning far sparser scenes than the data. Applied instead to iARCS output, the same filter retains the geometric gains with much less diversity loss.

\begin{table}[h]
\centering
\small
\setlength{\tabcolsep}{3pt}
\renewcommand{\arraystretch}{1.1}
\resizebox{\columnwidth}{!}{%
\begin{tabular}{llccccc}
\toprule
Room & Method & Draws & Obj. & $Col_{obj}\downarrow$ & FID$\downarrow$ & SCA \\
\midrule
\multirow{3}{*}{Bed}  & iARCS      & \textbf{1{,}080} & \textbf{5.12} & 40.45\% & \textbf{1.60} & 66.61\% \\
                      & Rej. Sampling  & 4{,}000 & 4.55 & 26.63\% & 1.84 & 76.56\% \\
                      & iARCS + Rej.  & 4{,}000 & 4.14 & \textbf{25.07\%} & 1.81 & \textbf{65.11\%} \\
\midrule
\multirow{3}{*}{Living} & iARCS    & \textbf{1{,}080} & \textbf{11.36} & 58.33\% & \textbf{3.10} & 84.12\% \\
                      & Rej. Sampling  & 4{,}000 & 7.20 & \textbf{56.09\%} & 5.05 & 88.96\% \\
                      & iARCS + Rej.  & 4{,}000 & 9.82 & 60.93\% & 4.17 & \textbf{80.25\%} \\
\midrule
\multirow{3}{*}{Dining} & iARCS    & \textbf{1{,}080} & \textbf{10.20} & 51.34\% & \textbf{2.65} & 78.74\% \\
                      & Rej. Sampling  & 4{,}000 & 4.41 & \textbf{48.81\%} & 9.70 & 94.56\% \\
                      & iARCS + Rej.  & 4{,}000 & 9.03 & 66.56\% & 5.27 & \textbf{77.35\%} \\
\bottomrule
\end{tabular}%
}
\caption{\textbf{Rejection sampling.} iARCS compared against rejection sampling
from the pretrained model, and against the same filter applied on top of iARCS,
for each room type.}
\label{tab:rejection}
\end{table}

\section{Evaluation Metric Definitions}
\label{app:metrics}

\noindent\textbf{Distribution metrics.} CLIP-FID is a Fr\'echet distance computed on CLIP embeddings~\cite{radford2021learningtransferablevisualmodels} of top-down scene projections, used in place of InceptionNet class predictions because our inputs are layout projections rather than natural images. Scene Classification Accuracy (SCA) is the accuracy of a classifier trained to separate real from synthesized scenes; values near the 50\% chance level indicate synthetic scenes the classifier cannot distinguish from real ones. Together these quantify perceptual fidelity, dataset coverage and diversity relative to the target distribution.

\noindent\textbf{Task success.} For each task, \emph{Success} is the fraction of generated scenes that satisfy the constraint, evaluated by a binary geometric check thresholded on the same constraint decomposition that defines the task reward. It is therefore the constraint applied as a hard predicate rather than as a shaped signal, and is by construction not independent of the training objective: the constraint \emph{is} the task. All methods are scored with the identical check, including the constraint-satisfying subset of real scenes (3D-FRONT$^{*}$) and the unsteered base model, so the comparison across rows remains matched. The metrics that are never optimized during training, and are therefore independent of it, are $R_{reach}$, $R_{walkable}$, CLIP-FID and SCA.

\noindent\textbf{Physical plausibility.} The object-collision rate $Col_{obj}$ is the percentage of objects colliding with any other object, and the scene-collision rate $Col_{scene}$ is the percentage of scenes containing at least one collision. A pair of objects counts as colliding if their 3D bounding-box IoU is strictly greater than zero, so any non-zero overlap volume is a collision.

\noindent\textbf{Floor-plan compliance and functional utility.} The out-of-bound placement rate $R_{out}$ is the fraction of objects placed outside the room boundary. The reachable-object ratio $R_{reach}$ is measured from sampled valid starting locations, and the walkability score $R_{walkable}$ is the ratio between the area of the largest connected walkable region and the total walkable area.

\section{Room-Wise Scene Synthesis Results}
\label{app:roomwise}

Table~\ref{tab:roomwise_scene_synthesis} reports the complete scene-synthesis results for bedrooms, living rooms, and dining rooms using the metrics defined above.

\begin{table*}[t]
\centering
\caption{All values are percentages except $R_{walkable}$ and CLIP-FID. Quantitative comparison of indoor scene synthesis methods on physical plausibility, functional utility, and distribution-level quality. Among the generated methods, iARCS achieves the best object-collision rate, boundary compliance, reachability, and walkability in all three room types while maintaining competitive realism and diversity.}
\label{tab:roomwise_scene_synthesis}
\setlength{\tabcolsep}{1.5pt}
\resizebox{\textwidth}{!}{%
\renewcommand{\arraystretch}{1.15}
\begin{tabular}{l|ccc|cc|cc|ccc|cc|cc|ccc|cc|cc}
\toprule
\multirow{3}{*}{Method} & \multicolumn{7}{c|}{Bedroom} & \multicolumn{7}{c|}{Living Room} & \multicolumn{7}{c}{Dining Room} \\
\cline{2-22}
& \multicolumn{3}{c|}{Physical} & \multicolumn{2}{c|}{Functional} & \multicolumn{2}{c|}{Diversity} & \multicolumn{3}{c|}{Physical} & \multicolumn{2}{c|}{Functional} & \multicolumn{2}{c|}{Diversity} & \multicolumn{3}{c|}{Physical} & \multicolumn{2}{c|}{Functional} & \multicolumn{2}{c}{Diversity} \\
\cline{2-22}
& $Col_{obj}\downarrow$ & $Col_{scene}\downarrow$ & $R_{out}\downarrow$ & $R_{reach}\uparrow$ & $R_{walkable}\uparrow$ & CLIP-FID$\downarrow$ & SCA$\downarrow$ & $Col_{obj}\downarrow$ & $Col_{scene}\downarrow$ & $R_{out}\downarrow$ & $R_{reach}\uparrow$ & $R_{walkable}\uparrow$ & CLIP-FID$\downarrow$ & SCA$\downarrow$ & $Col_{obj}\downarrow$ & $Col_{scene}\downarrow$ & $R_{out}\downarrow$ & $R_{reach}\uparrow$ & $R_{walkable}\uparrow$ & CLIP-FID$\downarrow$ & SCA$\downarrow$ \\
\midrule
Ground Truth (GT) & 42.00 & 72.04 & 5.79 & 82.80 & 0.841 & -- & -- & 39.93 & 79.00 & 1.49 & 71.86 & 0.882 & -- & -- & 47.43 & 89.20 & 0.73 & 59.87 & 0.829 & -- & -- \\
\midrule
ATISS & 72.39 & \textbf{53.61} & 16.35 & 60.27 & 0.839 & 1.52 & 79.24 & 67.81 & 95.35 & 30.12 & 67.23 & 0.792 & \textbf{3.05} & 89.12 & 75.12 & 98.10 & 13.10 & 68.23 & 0.812 & 2.82 & 80.12 \\
MiDiffusion & 52.67 & 81.67 & 5.89 & 85.70 & 0.806 & \textbf{1.40} & 65.99 & 64.97 & 93.10 & 31.90 & 75.42 & 0.856 & 3.26 & 85.92 & 71.31 & 99.70 & 14.55 & 70.92 & 0.824 & 2.83 & 79.96 \\
InstructScene & 43.48 & 74.07 & 37.98 & 80.14 & 0.798 & 5.23 & 80.29 & 63.21 & 95.83 & 46.07 & 63.67 & 0.885 & 5.69 & 87.73 & 52.17 & \textbf{94.22} & 46.11 & 63.86 & 0.846 & 5.88 & 83.88 \\
PhyScene & 45.35 & 76.04 & 20.32 & 74.72 & 0.813 & 1.49 & \textbf{64.55} & 60.62 & 93.92 & 29.85 & 72.14 & 0.871 & 3.36 & 86.81 & 55.64 & 94.62 & 22.70 & 64.79 & 0.864 & \textbf{2.60} & 84.51 \\
iARCS (Ours) & \textbf{40.45} & 64.63 & \textbf{2.94} & \textbf{87.82} & \textbf{0.861} & 1.60 & 66.61 & \textbf{58.33} & \textbf{65.50} & \textbf{20.79} & \textbf{80.17} & \textbf{0.955} & 3.10 & \textbf{84.12} & \textbf{51.34} & 94.50 & \textbf{9.23} & \textbf{73.18} & \textbf{0.873} & 2.65 & \textbf{78.74} \\
\bottomrule
\end{tabular}%
}
\end{table*}

\section{Additional Ablations on Iterative Reward Refinement}
\label{app:iterative}

\begin{figure*}[h]
\centering
\includegraphics[width=\textwidth]{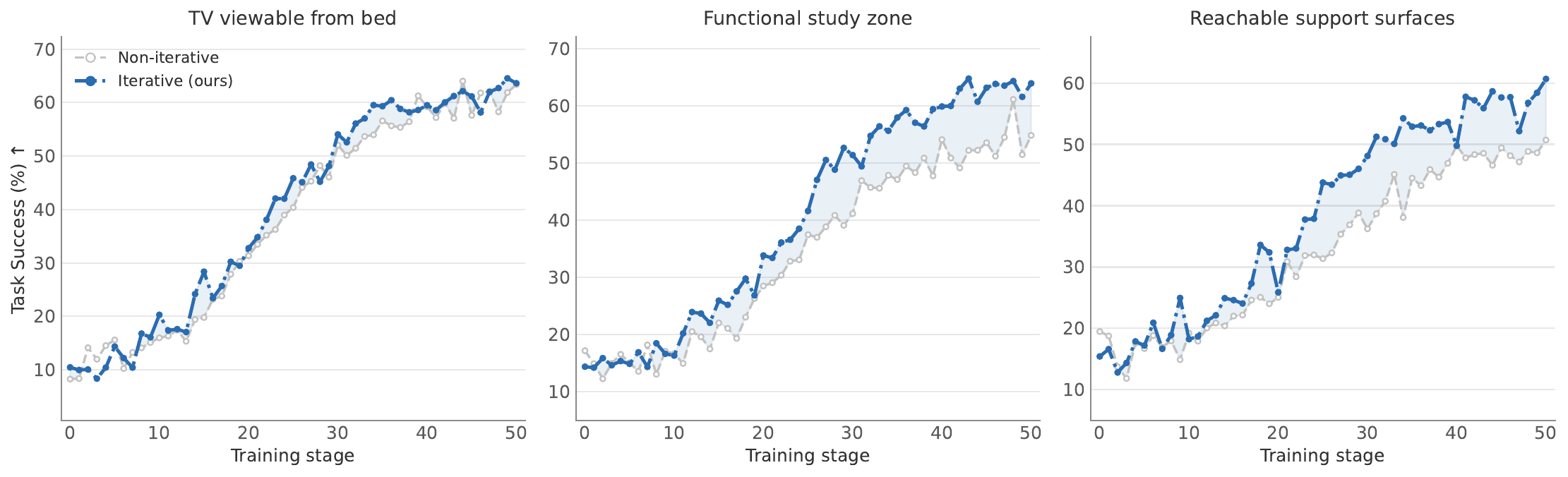}
\caption{\textbf{Effect of iterative reward refinement.} We compare task-specific RL training with and without iterative reward refinement across three constraints. While single-pass rewards can sometimes match iterative refinement, the iterative loop provides clear benefits when the initial reward is overly complex, too strict, or better suited to a curriculum.}
\label{fig:iterative_refinement_curves}
\end{figure*}

Figure~\ref{fig:iterative_refinement_curves} provides additional ablations comparing reward optimization with and without iterative reward refinement across three task-specific constraints. Some rewards can perform similarly without iteration; however, iterative refinement is substantially more meaningful for constraints where the initial LLM-generated reward is overly strict, poorly shaped, or too difficult for RL to optimize directly.

\begin{figure}[h]
\centering
\includegraphics[width=\columnwidth]{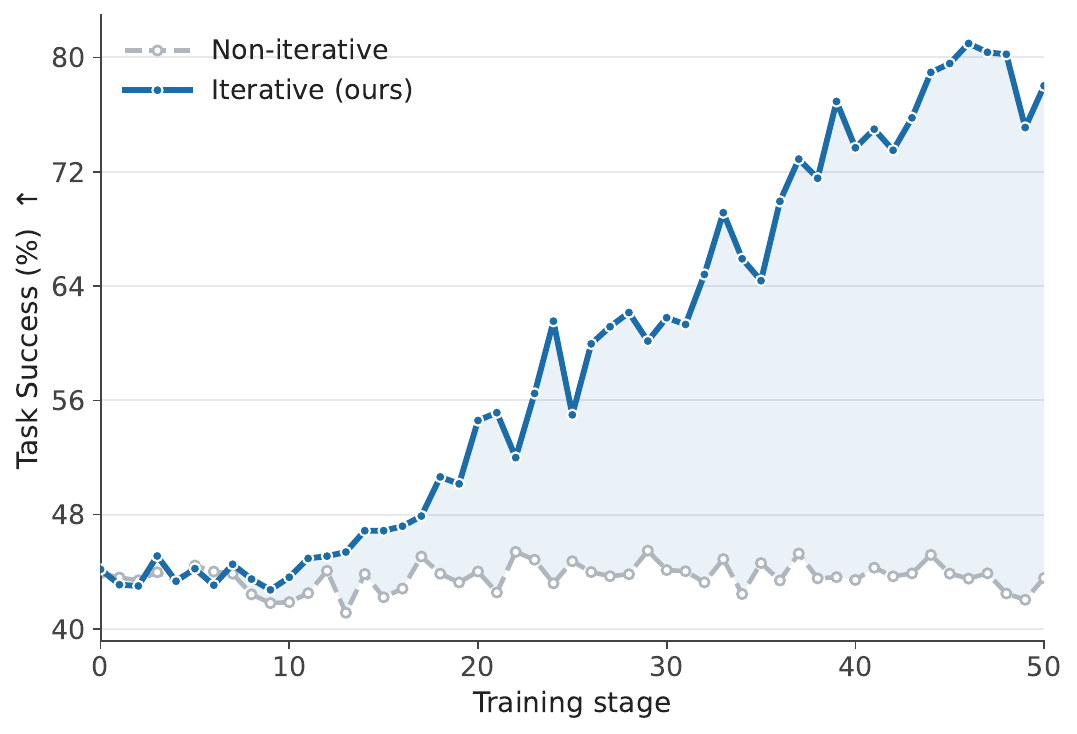}
\caption{\textbf{Effect of iterative reward refinement.} For the prompt ``A bedroom with 2 night tables on either side of the bed,'' iteration identifies violations, refines the reward, and improves task adherence over single-pass reward generation.}
\label{fig:iterative_reward_refinement}

\end{figure}

For example, for the task ``bedroom with two night tables on either side of the bed,'' the Gemini model initially produced an unnecessarily complicated reward with hard thresholds requiring both night tables to be equidistant and exactly 1~m from the bed. After observing that RL training could not reliably improve under this overly rigid objective, the subsequent iterations simplified the reward code to focus on the semantically important constraints: placing one night table on each side of the bed only. For the task ``all support surfaces less than 1~m,'' the model instead used curriculum learning: it first encouraged support objects to be below a relaxed height threshold of 1.5~m, then progressively decreased the target height toward the final 1~m constraint. These examples show that reward reflection is essential. Figure~\ref{fig:iterative_reward_refinement} illustrates the night-table case directly, comparing scenes generated with and without refinement.

\section{Iterative Agentic Reward System}
\label{app:agentic}

The agentic loop of Fig.~\ref{fig:overview} runs in three steps: constraint decomposition, reward-code generation, and reward reflection. Each step is one LLM call with a fixed system prompt. We describe what each prompt contains rather than reproducing it verbatim.

\noindent\textbf{Shared context.} All three prompts start from the same context block: (i) a description of 3D-FRONT and its conventions (the $Y$-axis is vertical and the $XZ$-plane is the floor, units are metres, ceiling objects sit near the ceiling height, floor objects are centred at half their height, empty slots carry the last class index, and compass directions run along $\pm X$ and $\pm Z$); (ii) per-room dataset statistics in JSON, such as class frequencies; and (iii) a description of the universal rewards already optimized in Phase~1, so that the agent does not re-derive them. Steps~2 and~3 also describe the batched scene tensors the reward code receives (object positions, half-extent sizes, class indices, an empty-slot mask, and orientation as $(\cos\theta,\sin\theta)$), together with a library of geometric utility functions the code may call.

\subsection{Step 1: Constraint Decomposition}

The LLM decomposes the user prompt into verifiable constraints. Example prompts include:
\begin{itemize}
    \item ``Robot grasping scene where all support surfaces are within a 1.0 m vertical reach limit''
    \item ``Room with a TV stand positioned so a farsighted person can view a 4K TV from bed''
    \item ``A bedroom scene with a functional study zone''
\end{itemize}
The response is a JSON list of constraints, each with an identifier, a snake-case name and a natural-language description, for example:
\begin{lstlisting}[language=json, basicstyle=\scriptsize\ttfamily]
{
  "constraints": [
    {"id": "C1", "name": "tv_stand_present",
     "description": "Scene contains TV stand"},
    {"id": "C2", "name": "tv_viewable_from_bed",
     "description": "TV within 2-4m of bed"}
  ]
}
\end{lstlisting}

\subsection{Step 2: Reward Code Generation}

For each constraint $C_i$, the LLM writes a Python reward function $R_i$ following this template:
\begin{lstlisting}[language=Python, basicstyle=\scriptsize\ttfamily]
def get_reward(parsed_scenes, idx_to_labels,
               room_type, floor_polygons, **kwargs):
    # Compute reward for constraint
    # Returns: torch.Tensor of shape (B,)
    return rewards

def test_reward(...):
    # Test with multiple scenes
    assert <TEST_CASE_1>
    assert <TEST_CASE_2>
\end{lstlisting}
The prompt requires every function to provide (1) a \textbf{success threshold} in raw reward units, above which the constraint counts as satisfied; (2) \textbf{bounded rewards}, so that anomalous scenes receive a worse but capped value and never an infinite one; and (3) \textbf{test cases}: \texttt{test\_reward} builds a few small synthetic scenes with a helper function and asserts the expected rewards, printing expected and actual values on failure so that reward scale can be corrected in the next iteration. The response is JSON giving, for each constraint, the reward code and its success threshold.

\subsection{Step 3: Reward Reflection}

At each reflection step the agent receives the current reward program, the reward statistics summarised below, and the evolution memory of previously generated programs. Top-down views of 10 scenes sampled from the current policy are rendered and passed alongside these statistics.

The reflection prompt adds the user prompt, the current constraints and reward programs, and per-reward statistics computed on both the 3D-FRONT room split (4,042 bedroom scenes) and scenes generated by the current model (1,080 scenes): success rate, mean, median, range, standard deviation, percentiles from P1 to P99, skewness, kurtosis, and the fraction of scenes at or near the maximum and minimum reward. The agent is asked to plan a curriculum and to return, as JSON, the curriculum stages together with constraints and reward programs for the current stage only.

\noindent\textbf{Curriculum Strategy.} We use iterative RL:
\begin{enumerate}
    \item \textbf{Iteration 1}: Focus on 2-3 critical constraints with low baseline success
    \item \textbf{Analysis}: Evaluate 1,080 new scenes, compare success rates
    \item \textbf{Iteration $n+1$}: Retain learned constraints, add new ones, refine if needed
\end{enumerate}

\section{Reward Synthesis Example}
\label{app:reward_example}

\begin{figure*}[h]
    \centering
    \includegraphics[width=0.8\textwidth]{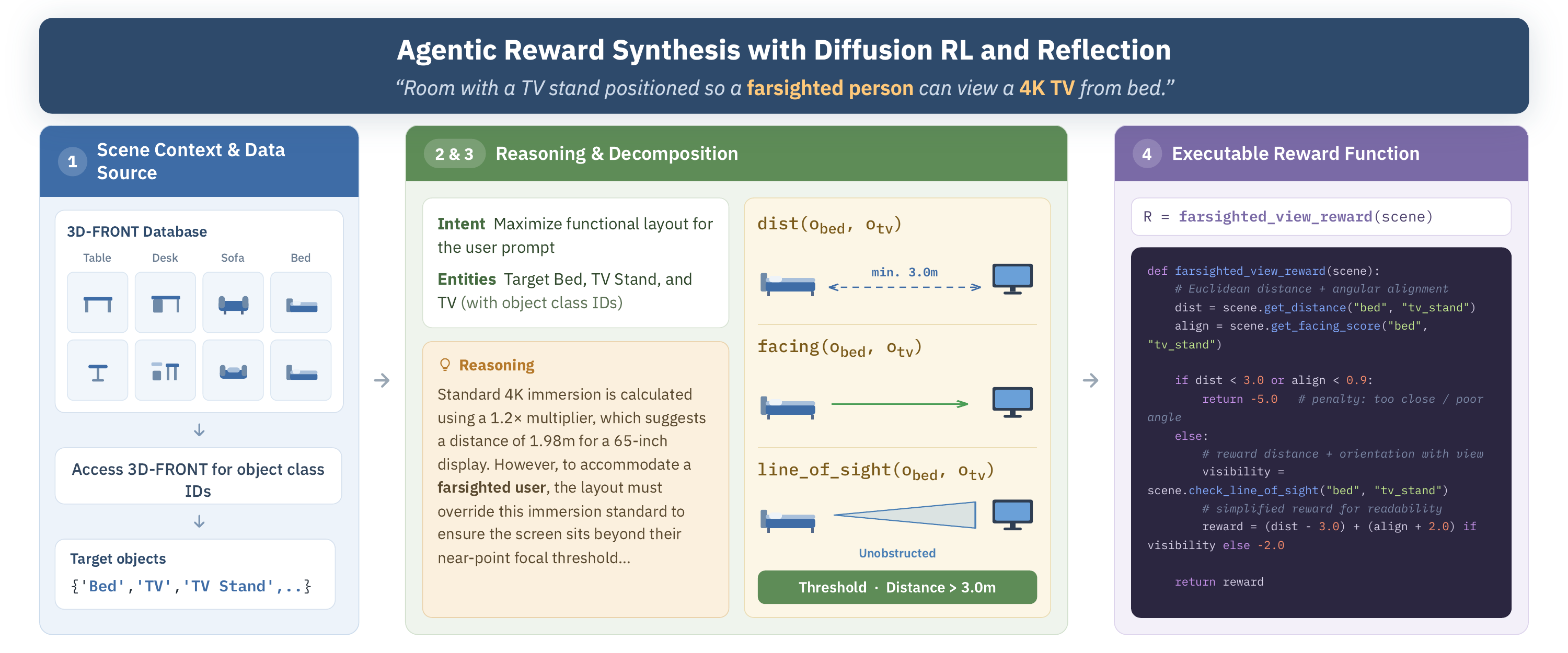}
    \caption{\textbf{Example of agentic reward synthesis for a task-specific prompt.} Given the prompt ``Room with a TV stand positioned so a farsighted person can view a 4K TV from bed,'' the LLM pipeline retrieves scene context from 3D-FRONT objects, decomposes intent into geometric checks such as viewing distance and orientation, and outputs executable reward code for diffusion RL.}
    \label{fig:reward_function_qualitative}
\end{figure*}

Figure~\ref{fig:reward_function_qualitative} illustrates how a natural-language task prompt is converted into measurable geometric checks and executable reward code used by the diffusion RL loop.

\section{Qualitative Samples}
\label{app:qualitative}

\begin{figure*}[h]
\centering
\includegraphics[width=\textwidth, trim=0 175 0 0, clip]{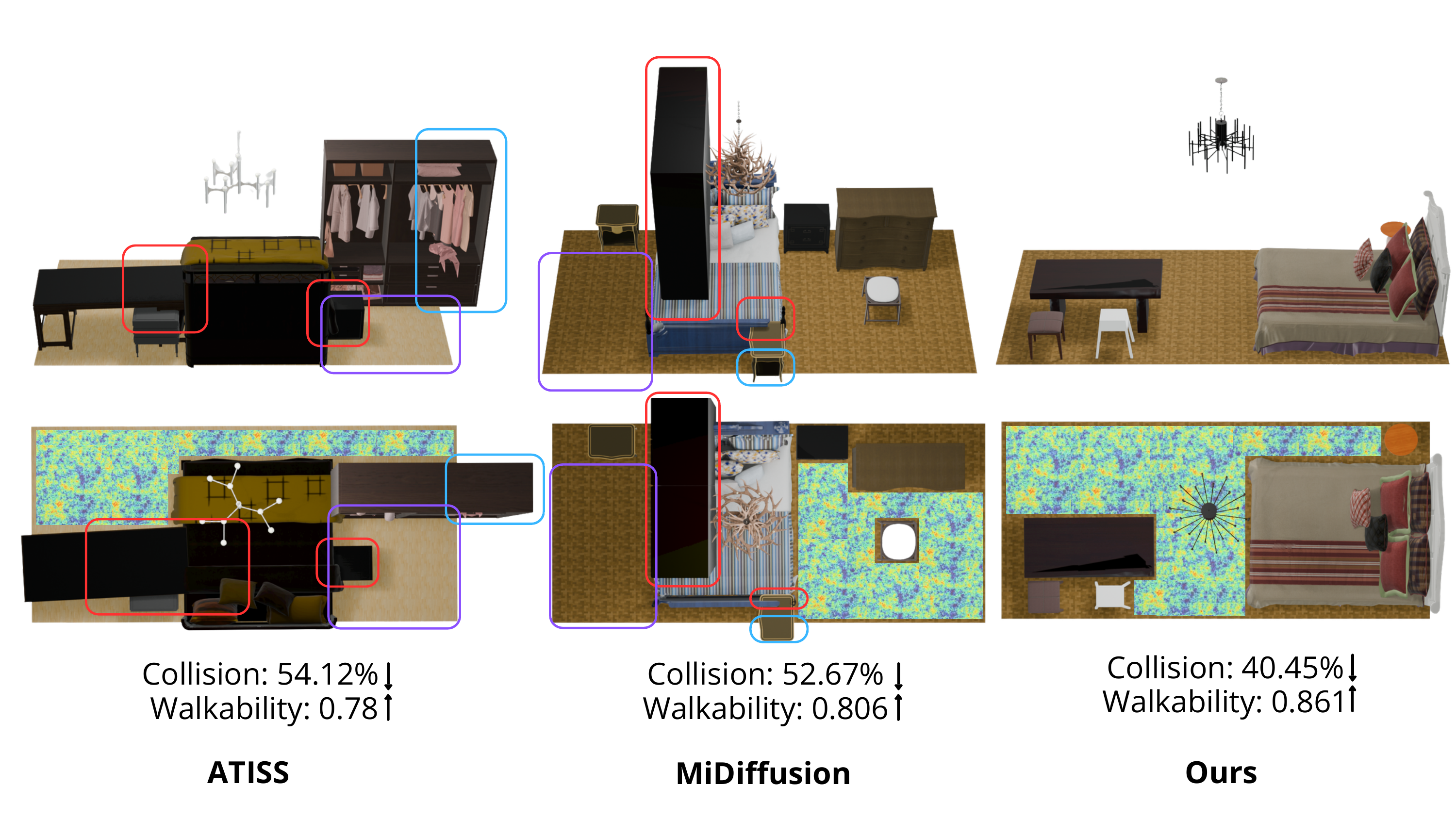}
\caption{Qualitative sample comparison. Left to right: ATISS, MiDiffusion, and iARCS (ours).}
\label{fig:qualitative_supp1}
\end{figure*}

\begin{figure*}[h]
\centering
\includegraphics[width=0.7\textwidth]{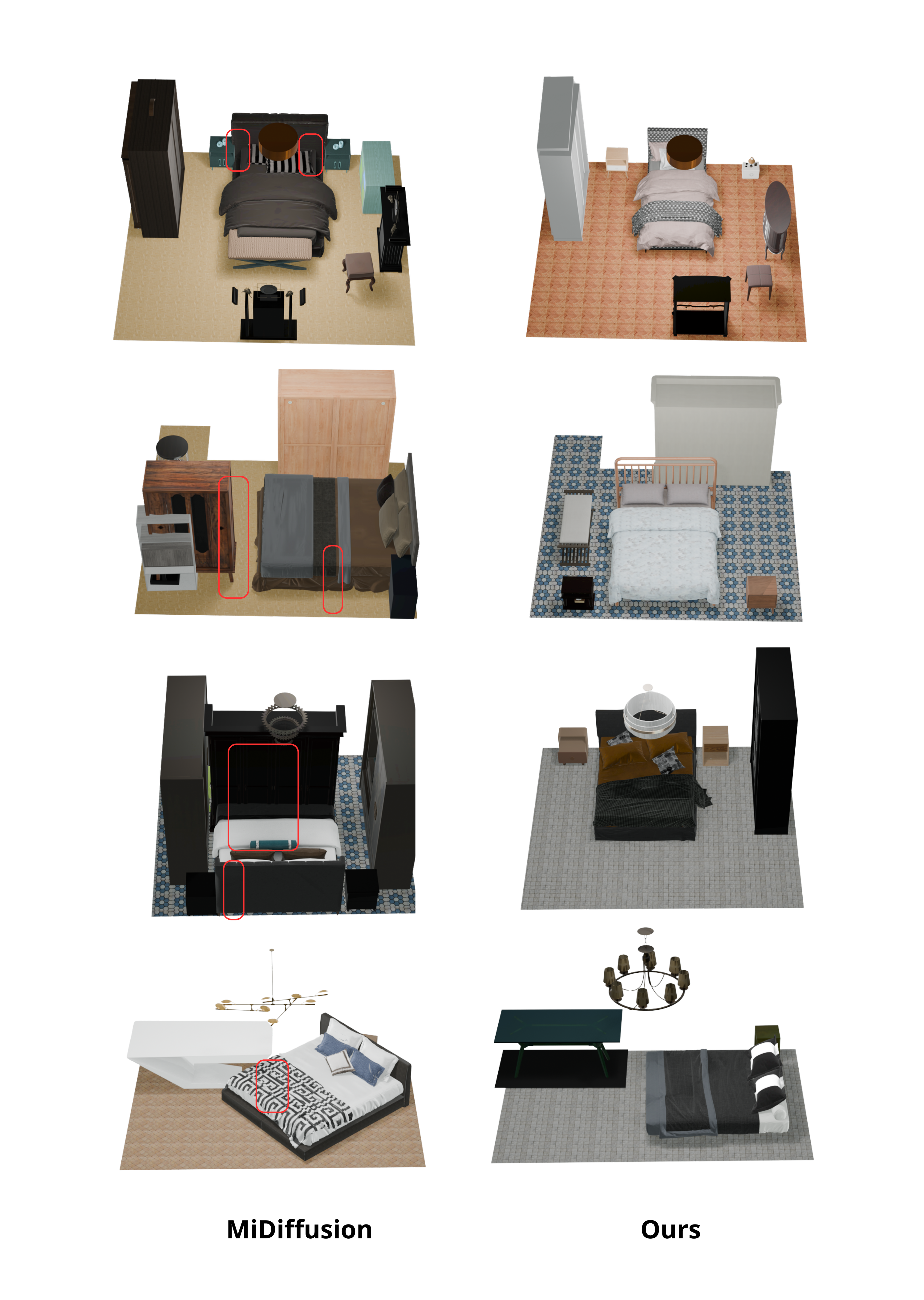}
\caption{Qualitative sample comparison of MiDiffusion and iARCS.}
\label{fig:qualitative_supp2}
\end{figure*}

Figure~\ref{fig:qualitative_supp1} and Figure~\ref{fig:qualitative_supp2} show rendered scene comparisons across methods. Our method (iARCS) produces more physically plausible scenes than baseline models, with fewer collisions, better object placement, and improved navigable free space.

\end{document}